%% file: neurips_2019.tex
\documentclass{article}

\usepackage[final]{neurips_2019}

\usepackage[utf8]{inputenc} % allow utf-8 input
\usepackage[T1]{fontenc}    % use 8-bit T1 fonts
\usepackage{hyperref}       % hyperlinks
\usepackage{url}            % simple URL typesetting
\usepackage{booktabs}       % professional-quality tables
\usepackage{amsfonts}       % blackboard math symbols
\usepackage{nicefrac}       % compact symbols for 1/2, etc.
\usepackage{microtype}      % microtypography

\usepackage{junstyle}
\usepackage{wrapfig}
\usepackage{float}
\usepackage{amssymb,amsmath}
\usepackage{bbm}
\usepackage{tabularx}
\usepackage{multirow}
\usepackage{array}
\usepackage{natbib}
\input{tex/math_commands.tex}

\title{Deep Generative Video Compression}

\author{Jun Han\thanks{~~Shared first authorship.}\\
Dartmouth College\\ 
\texttt{\footnotesize  junhan@cs.dartmouth.edu}
\And Salvator Lombardo\footnotemark[1]\\
Disney Research LA \\
\texttt{\footnotesize salvator.d.lombardo@disney.com}
\AND
Christopher Schroers\\
DisneyResearch|Studios\\
\texttt{\footnotesize  christopher.schroers@disney.com}
\And
Stephan Mandt\\
University of California, Irvine\\
\texttt{\footnotesize mandt@uci.edu}
}

\begin{document}

\maketitle

\begin{abstract}
The usage of deep generative models for image compression has led to impressive
performance gains over classical codecs while neural video compression is still in its infancy.
Here, we propose an end-to-end, deep generative modeling approach to compress temporal sequences with a focus on video. Our approach builds upon variational autoencoder (VAE) models for sequential data and combines them with recent work on neural image compression. The approach jointly learns to transform the original sequence into a lower-dimensional representation as well as to discretize and entropy code this representation according to predictions of the sequential VAE.  Rate-distortion evaluations on small videos from public data sets with varying complexity and diversity show that our model yields competitive results when trained on generic video content. Extreme compression performance is achieved when training the model on specialized content. 
\end{abstract}

\section{Introduction}

The transmission of video content is responsible for up to 80\% of the consumer internet traffic, and both the overall internet traffic as well as the share of video data is expected to increase even further in the future \citep{cisco2017forecast}. Improving compression efficiency is more crucial than ever. The most commonly used standard is H.264 \citep{wiegand2003overview}; more recent codecs include H.265 \citep{sullivan2012overview} %, also known as High Efficiency Video Coding (HEVC), 
and VP9 \citep{mukherjee2015technical}. 
All of these existing codecs follow the same block based hybrid structure \citep{musmann1985advances} which essentially emerged from engineering out and refining this concept over decades. From a high level perspective, they differ in a huge number of smaller design choices and have grown to become more and more complex systems.

While there is room for improving the block based hybrid approach even further \citep{VVC}, the question remains as to how much longer significant improvements can be obtained while following the same paradigm. In the context of image compression, deep learning approaches that are fundamentally different to existing codecs have already shown promising results \citep{balle2018variational, balle2016end, theis2017lossy, agustsson2017soft, minnen2018joint}.
Motivated by these successes for images, we propose a first step towards innovating beyond block-based hybrid codecs by framing video compression in a deep generative modeling context. To this end, we propose an unsupervised deep learning approach to encoding video. The approach simultaneously learns the optimal transformation of the video to a lower-dimensional representation \textit{and} a powerful predictive model that assigns probabilities to video segments, allowing us to efficiently entropy-code the discretized latent representation into a short code length.

Our end-to-end neural video compression scheme is based on sequential variational autoencoders~\citep{bayer2014learning,chung2015recurrent,li2018deep}.
%and the approach of \citet{balle2016end} for discretizing and entropy coding a continuous latent representation.
The transformations to and from the latent representation (the encoder and decoder) are parametrized by deep neural networks and are learned by unsupervised training on videos. These latent states have to be discretized before they can be compressed into binary. \citet{balle2016end} address this problem by using a box-shaped variational distribution with a fixed width, forcing the VAE to `forget’ all information stored on smaller length scales due to the insertion of noise during training. This paper follows the same paradigm for temporally-conditioned distributions. A sequence of quantized latent representations still contains redundant information as the latents are highly correlated. (Lossless) entropy encoding exploits this fact to further reduce the expected file size by expressing likely data in fewer bits and unlikely data in more bits. This requires knowledge of the probability distribution over the discretized data that is to be compressed, which our approach obtains from the sequential prior.

Among the many architectural choices that our approach enables, we empirically investigate a model that is well suited for the regime of extreme compression. This model uses a combination of both \emph{local} latent variables, which are inferred from a single frame, and a \emph{global} state, inferred from a multi-frame segment, to efficiently store a video sequence. The dynamics of the local latent variables are modeled stochastically by a deep generative model. After training, the context-dependent predictive model is used to entropy code the latent variables into binary with an arithmetic coder.

\begin{figure}[t]
\label{fig:sprite}
\begin{center}
\begingroup
\setlength{\tabcolsep}{1pt} % Default value: 6pt
\begin{tabular}{cc}
 \multicolumn{2}{c}{{\small H.265 (\textbf{21.1} dB @ \textbf{0.86} bpp)}} \\
\includegraphics[width=.15\textwidth]{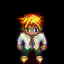}&
\includegraphics[width=.15\textwidth]{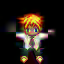} \\%[5pt]
%{\small t=1} & {\small t=6} \\
\end{tabular}
\begin{tabular}{cc}
 \multicolumn{2}{c}{{\small VP9 (\textbf{26.0} dB @ \textbf{0.57} bpp)}} \\
\includegraphics[width=.15\textwidth]{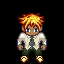}&
\includegraphics[width=.15\textwidth]{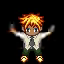} \\%[5pt]
%{\small t=1} & {\small t=6} \\
\end{tabular}
\begin{tabular}{cc}
 \multicolumn{2}{c}{{\small Ours (\textbf{44.6} dB @ \textbf{0.06} bpp)}} \\
% {\small Ours} & {\small H.265} & {\small VP9}\\
\includegraphics[width=.15\textwidth]{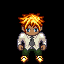}&
\includegraphics[width=.15\textwidth]{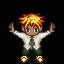} \\%[5pt]
%{\small t=1} & {\small t=6} \\
\end{tabular}
\endgroup 
\caption{
Reconstructed video frames using the established codecs H.265 (left), VP9 (middle), and ours (right), with videos taken from the Sprites data set (Section~\ref{sec:experiments}). On specialized content as shown here,  higher PSNR values in dB (corresponding to lower distortion) can be achieved at almost an order of magnitude smaller bits per pixel (bpp) rates. Compared to the classical codecs, fewer geometrical artifacts are apparent in our approach. 
%By training our method on specialized content data sets, it can achieve a significantly higher PSNR (lower distortion) even when run at a fraction of the bit rate of existing codecs.
} 
\end{center}
\end{figure}

In this paper, we focus on low-resolution video ($64\times64$) 
as the first step towards deep generative video compression. 
Figure \ref{fig:sprite} shows a test example of the possible performance improvements using our approach if the model is trained on restricted content (video game characters). The plots show two frames of a video, compressed and reconstructed by our approach and by classical video codecs. One sees that fine granular details, such as the hands of the cartoon character, are lost in the classical approach due to artifacts from block motion estimation (low bitrate regime), whereas our deep learning approach successfully captures these details with less than 10\% of the file length.

Our contributions are as follows:

\textbf{1)} \textbf{A general paradigm for generative compression of sequential data.} We propose a general framework for compressing sequential data by employing a sequential variational autoencoder (VAE) in conjuction with discretization and entropy coding to build an end-to-end trainable codec. 

\textbf{2)} \textbf{A new neural codec for video compression.} We employ the above paradigm towards building an end-to-end trainable codec. To the best of our knowledge, this is the first work to utilize a deep generative video model together with discretization and entropy coding to perform video compression. 

%\textbf{2)} Global inference. Temporal redundancy in a video can be taken into account by a temporal prior on each frame or through an architectural design that encodes an entire video segment into a global state. We propose a model which incorporates both global and local latent variables and show that this produces a shorter code length for a given image quality. 

\textbf{3)}  \textbf{High compression ratios.} We perform experiments on three public data sets of varying complexity and diversity. Performance is evaluated in terms of rate-distortion curves. For the low-resolution videos considered in this paper, our method is competitive with traditional codecs after training and testing on a diverse set of videos. Extreme compression performance can be achieved on a restricted set of videos containing specialized content if the model is trained on similar videos. 

\textbf{4)} \textbf{Efficient compression from a global state.} While a deep latent time series model takes temporal redundancies in the video into account, one optional variation of our model architecture tries to compress static information into a separate global variable~\citep{li2018deep} which acts similarly as a key frame in traditional methods. We show that this decomposition can be beneficial.
%leading to an approximately disentangledv representation~\citep{li2018deep}. We show that this decomposition can be beneficial.

Our paper is organized as follows. In Section~\ref{sec:related},  we summarize related works before describing our method in Section~\ref{sec:mainmethod}. Section~\ref{sec:experiments} discusses our experimental results. We give our conclusions in Section~\ref{sec:conclusions}.

\section{Related Work} \label{sec:related}
The approaches related to our method fall into three categories: deep generative video models, neural image compression, and neural video compression.

\paragraph{Deep generative video models.} Several works have applied the variational autoencoder (VAE)~\citep{kingma2013auto,rezende2014stochastic} to stochastically model sequences~\citep{bayer2014learning,chung2015recurrent}.  \citet{babaeizadeh2017stochastic, xu2018stochastic} use a VAE for stochastic video generation. \citet{he2018probabilistic} and \citet{denton2018stochastic} apply a long short term memory (LSTM) in conjunction with a sequential VAE to model the evolution of the latent space across many video frames. \citet{li2018deep} separate latent variables of a sequential VAE into local and global variables in order to learn a disentangled representation for video generation. 
\citet{vondrick2016generating} generate realistic videos by using a generative adversarial network \citep{goodfellow2014generative} to learn to separate foreground and background, and \citet{lee2018stochastic} combine variational and adversarial methods to generate realistic videos. 
This paper also employs a deep generative model to model the sequential probability distribution of frames from a video source.
In contrast to other work, our method learns a continuous latent representation that can be discretized with minimal information loss, required for further compression into binary. Furthermore, our objective is to convert the original video into a short binary description rather than to generate new videos.

\paragraph{Neural image compression.} There has been significant work on applying deep learning to image compression. In \citet{toderici2015variable,toderici2017full,johnston2018improved}, an LSTM based codec is used to model spatial correlations of pixel values and can achieve different bit-rates without having to retrain the model. 
\citet{balle2016end} perform image compression with a VAE and demonstrate how to approximately discretize the VAE latent space by introducing noise during training. This work is refined by \citep{balle2018variational} which improves the prior model (used for entropy coding) beyond the mean-field approximation by transmitting side information in the form of a hierarchical model. \citet{minnen2018joint} consider an autoregressive model to achieve a similar effect. \cite{santurkar2018generative} studies the performance of generative compression on images and suggests it may be more resilient to bit error rates. 
These image codecs encode each image independently and therefore their probabilistic models are stationary with respect to time. In contrast, our method performs compression according to a non-stationary, time-dependent probability model which typically has lower entropy per pixel.

\paragraph{Neural video compression.} The use of deep neural networks for video compression is relatively new. \citet{wu2018video} perform video compression through image interpolation between reference frames using a predictive model based on a deep neural network. \cite{chen2017deepcoder} and \cite{chen2018learning} use a deep neural architecture to predict the most likely frame with a modified form of block motion prediction and store residuals in a lossy representation. Since these works are based on motion estimation and residuals, they are somewhat similar in function and performance to existing codecs. \cite{lu2019dvc} and \cite{djelouah2019neuralinterframe} also follow a pipeline based on motion estimation and residual computation as in existing codecs. In contrast,  our method is not based on motion estimation, and the full inferred probability distribution over the space of plausible subsequent frames is used for entropy coding the frame sequence (rather than residuals). In a concurrent publication, \cite{habibian2019video} perform video compression by utilizing a 3D variational autoencoder. In this case, the 3D encoder removes temporal redundancy by decorrelating latents, wheras our method uses entropy coding (with time-dependent probabilities) to remove temporal redundancy.

\section{Deep Generative Video Compression}
% or again Deep probabilistic video compression
\label{sec:mainmethod}
%\label{sec:overview}

\newcommand{\rate}{\mathcal{R}}
\newcommand{\distortion}{\mathcal{D}}
\newcommand{\enc}{\varphi}
\newcommand{\dec}{\psi}

\begin{figure*}[t]
\center
\includegraphics[width=.85\textwidth]{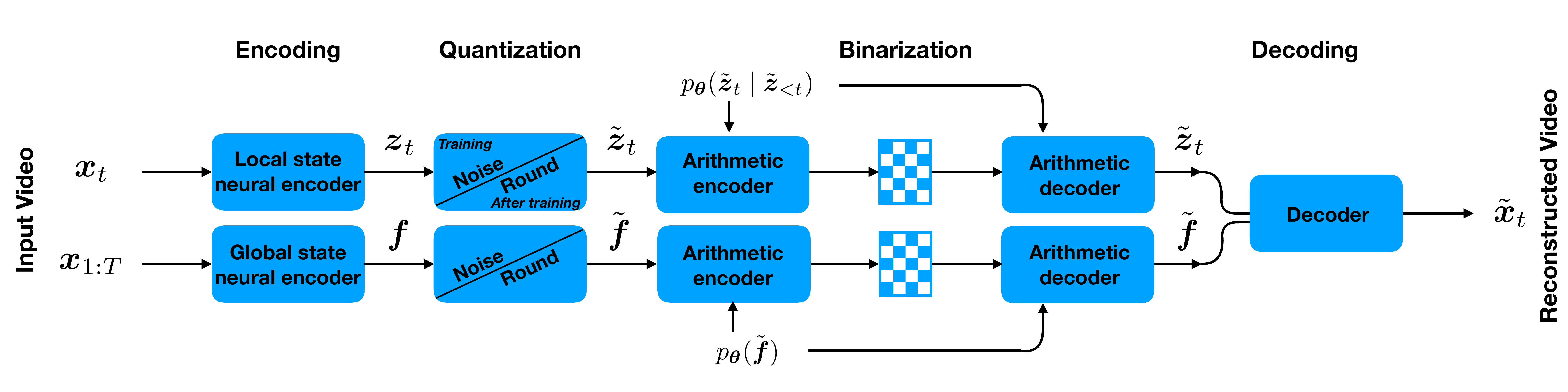}
% \includegraphics[width=.95\textwidth]{figures/encoder-orig.pdf}
% \caption{Operational diagram of our compression codec. A video segment is encoded into per-frame latent variables $\vz_t$ and per-segment global state $\bd{f}$, which are then quantized and arithmetically encoded into binary according to the prior model. To recover an approximation to the original video, the latent variables are arithmetically decoded from the binary and passed through the decoder.}
\caption{High-level operational diagram of our compression codec (see Section 3). A video segment is encoded into per-frame latent variables $\vz_t$ and (optionally) also into a per-segment global state $\bd{f}$ using a VAE architecture. Both latent variables are then quantized and arithmetically encoded into binary according to the respective prior models. To recover an approximation to the original video, the latent variables are arithmetically decoded from the binary and passed through the neural decoder. 
%Both the encoding and the decoding processes are sequential, as the prior model (used for arithmetic coding) conditions on latent states from previous frames.
}
\label{fig:diagram}
\end{figure*}

Our end-to-end approach simultaneously learns to transform a video into a lower-dimensional latent representation \emph{and} to remove the remaining redundancy in the latents through model-based entropy coding. Section~\ref{sec:main_overview} gives an overview of the deep generative video coding approach as a whole before Sections~\ref{sec:main_lossless} and~\ref{sec:main_lossy} detail on the model-based entropy coding and the lower-dimensional representation, respectively.

\subsection{Overview}
\label{sec:main_overview}

Lossy video compression is a constrained optimization problem that can be approached from two different angles: 1) either as finding the shortest description of a video without exceeding a certain level of information loss or 2) as finding the minimal level of information loss without exceeding a certain description length. Both optimization problems are equivalent with either a focus on description length (rate) or information loss (distortion) constraints. The distortion is a measure of how much error encoding and subsequent decoding incurs while the rate quantifies the amount of bits the encoded representation occupies. When denoting distortion by $\distortion$, rate by $\rate$, and the maximal rate constraint by $\rate_c$, the compression problem can be expressed as
\begin{equation*}
%\label{eqn:rd_constrained}
    \min  \distortion \quad \textrm{subject to}  \quad \rate \leq \rate_c.
\end{equation*}
Such a constrained formulation is often cumbersome but can be  solved in a Lagrange multiplier formulation, where
%~\citep{http://ip.hhi.de/imagecom_G1/assets/pdfs/rateDist_98.pdf-evtl.22or23therein}.
the rate and distortion terms are weighted against each other by a Lagrange multiplier $\beta$:
\begin{equation}
\label{eqn:rd_lagrangian}
    \min \distortion + \beta \rate.
\end{equation}
%
%Solving \eqref{eqn:rd_lagrangian} for a specific $\lambda$ is equal to solving \eqref{eqn:rd_constrained} for a particular rate constraint $\rate_c$~\citep{todo-previous-cite-or-cite-therein?}.

In existing video codecs, encoders and decoders have been meticulously engineered to improve coding efficiency.
Instead of engineering encoding and decoding functions, in our end-to-end machine learning approach we aim to \emph{learn} these mappings by parametrizing them by deep neural networks and then optimizing Eq.~\ref{eqn:rd_lagrangian} accordingly. 

There is a well-known equivalence~\citep{balle2018variational,alemi2018fixing} between the evidence lower bound in amortized variational inference~\citep{gershman2014amortized, zhang2017advances}, and the Lagrangian formulation of lossy coding of Eq.~\ref{eqn:rd_lagrangian}. Variational inference involves a probabilistic model $p(\vx, \vz) = p(\vx |\vz) p(\vz)$ over data $\vx$ and latent variables $\vz$. The goal is to lower-bound the marginal likelihood $p(\vx)$ using a variational distribution $q(\vz|\vx)$. When the variational distribution $q$ has a fixed entropy (e.g., by fixing its variance), this bound is, up to a constant,
\begin{equation}
\label{eq:generic_ELBO}
    \mathbb{E}_q [\log p(\vx|\vz)] - \beta \, H[q(\vz|\vx), p(\vz)],
\end{equation}
where $H$ is the cross entropy between the approximate posterior and the prior. When allowing for arbitrary $\beta$, \citet{balle2016end} showed in the context of image compression with variational autoencoders that the negative of Eq.~\ref{eq:generic_ELBO} becomes a manifestation of Eq.~\ref{eqn:rd_lagrangian}. While the first term measures the expected reconstruction error of the encoded images, 
 the cross entropy term becomes the expected code length as the (learned) prior $p(\vz)$ is used to inform a lossless entropy coder about the probabilities of the discretized encoded images.
 %\citet{balle2016end} proposed to employ this variational autoencoder approach to image compression. They used a box-shaped variational distribution, rounded the latent state to the nearest integer, and informed an arithmetic coder about the frequencies of integer-valued latent states using the prior. In this approach, the cross-entropy term becomes the expected code lengh.
% 
%After compressing an image into a latent state using a variational autoencoder trained to maximimize Eq.~\ref{eq:generic_ELBO}, the latent state is rounded to the nearest integer and compressed losslessly using arithmetic coding with propabilities given by the prior model. 
In this paper we generalize this approach to videos by employing probabilistic deep sequential latent state models.

Fig.~\ref{fig:diagram} summarizes our overall design. Given a sequence of frames $\vx_{1:T} = (\vx_1,\dots,\vx_T)$, we transform them into a sequence of latent states $\bm{z}_{1:T}$ and optionally also a global state $\bm{f}$.
%
%\CS{It could be interesting to mention briefly how connected encoding into f and $z_t$s is at training/testing time and how the "data flow" is more precisely. Can we just encode f separately and then each z sequentially? If it does not fit here it could also be mentioned elsewhere.}
%By separating local and global information in the video we allow the codec to store information that is common to the sequence of frames \emph{and} additional dynamical content. 
Although this transformation into a latent representation is lossy, the video is not yet optimally compressed as there are still correlations in the latent space variables. To remove this redundancy, the latent space must be entropy coded into binary. This is the distinguishing element between variational autoencoders and full compression algorithms. 
% Entropy coding requires a discrete latent space which we obtain by quantization. Furthermore, it requires a predictive model for latent space variables to convert them into a compact bit stream. 
 The bit stream can then be sent to a receiver where it is decoded into video frames. Our end-to-end machine learning approach simultaneously learns the predictive model required for entropy coding \emph{and} the optimal lossy transformation into the latent space. Both components are described in detail in the next sections, respectively.

\subsection{Entropy Coding via a Deep Sequential Model}
\label{sec:main_lossless}

Predictive modeling is crucial at the entropy coding stage. A better model which more accurately captures the true certainty about the next symbol has a smaller cross entropy with the data distribution and thus produces a bit rate that is closer to the theoretical lower bound for long sequences \citep{shannonmathematical}. For videos, temporal modeling is most important, making a learned temporal model an integral part of our model design. We now discuss a preliminary version of our model which does not yet include the global state, saving the specific details and encoder of our proposed model for Section~\ref{sec:main_lossy}. 

\paragraph{General model design.} 
When it comes to designing a generative model, the challenge over image compression is that videos exhibit strong temporal correlations in addition to the spatial correlations present in images. 
%A naive approach to neural video compression would be to encode the video frame-by-frame, using the marginal distribution of images as done in VAE image compression.
%However, this distribution does not capture the temporal correlations and therefore tends to have high entropy, leading to long code lengths. 
Treating a video segment as an independent data point in the latent representation (as would a 3D autoencoder) leads to data sparseness and poor generalization performance. Therefore, we propose to learn a temporally-conditioned prior distribution parametrized by a deep generative model to efficiently code the latent variables associated with each frame. 
Let $\vx_{1:T}=(\vx_1, \cdots, \vx_T)$ be the video sequence and $\vz_{1:T}$ be the associated latent variables. A generic generative model of this type takes the form: 
\begin{equation}
  p_\vthe(\vx_{1:T}, \vz_{1:T}) = \prod_{t=1}^T p_\vthe(\vz_t | \vz_{<t}) p_\vthe(\vx_t\mid \vz_t). 
\end{equation}
Above, $\vthe$ is shorthand for parameters of the model. By conditioning on previous frame latents in the sequence, the prior model can be more certain about the next $\vz_t$, thus achieving a smaller entropy and code length (after entropy coding). 

\paragraph{Arithmetic coding.}
 Entropy coding schemes require a discrete vocabulary, which is obtained in our case by rounding the latent states to the nearest integer after training. Care must be taken such that the quantization at inference time is approximated in a differentiable way during training. In practice, this is handled by introducing noise in the inference process. Besides dealing with quantization, we also need an accurate estimate of the probability density over the latent atoms for efficient coding. Knowledge of the sequential probability distribution of latents allows the entropy coder to decorrelate the bitstream so that the maximal amount of information per bit is stored~\citep{mackay2003information}. We obtain this probability estimation from the learned prior.

We employ an arithmetic coder~\citep{rissanen1979arithmetic, langdon1984introduction} to losslessly code the rounded latent variables into binary. In contrast to other forms of entropy encoding, such as Huffman coding,  arithmetic coding encodes the entire  sequence of discretized latent states $\vz_{1:T}$ into a single number. During encoding, the approach uses the conditional probabilities $p(\vz_t|\vz_{<t})$ to iteratively refine the real number interval $[0, 1)$ into a progressively smaller interval. After the sequence has been processed and a final (very small) interval is obtained, a binarized floating point number from the final interval is stored to encode the entire sequence of latents. Decoding the  decimal can similarly be performed iteratively by undoing the sequence of interval refinements to recover the original latent sequence. The fact that decoding happens in the same temporal order as encoding guarantees access to all conditional probabilities $p(\vz_t|\vz_{<t})$. Since $\vz_t$ was quantized, all probabilities for encoding and decoding exactly match.  In practice, besides iterating over time stamps $t$, we also iterate over the dimensions of $\vz_t$ during arithmetic coding.

\subsection{Proposed Generative and Inference Model}
\label{sec:main_lossy}

%We have described how model-based lossless entropy coding is performed. 
In this section, we describe the modeling aspects of our approach in more detail.  We refine the generative model to also include a global state which can be omitted to capture the base case outlined before. Besides the local state, the global state may help the model capture long-term information.  
%Specifically, we focus on the lossy transformation to the latent representation that reduces the dimension of the original input video.

%As follows, we describe the model-based lossy compression approach that reduces the dimension of the original input video. 
%We use a sequential variational autoencoder architecture similar to that of \citet{li2018deep}. An important departure from this architecture will be the inference model, whose noise distribution is not learned but fixed. This allows a better subsequent discretization. 

\paragraph{Decoder.} The decoder is a probabilistic neural network
that models the data as a function of their underlying latent codes. We use a stochastic recurrent variational autoencoder that transforms a sequence of local latent variables $\vz_{1:T}$ and a global state $\bd{f}$ into the frame sequence $\vx_{1:T}$, expressed by the following joint distribution:
%We propose a stochastic recurrent variational autoencoder to transform the video into a compressed representation of local latent variables $\vz_{1:T}$ and a global state $\bd{f}$, resulting in the following probabilistic deep generative model:
%
\begin{equation}
  p_\vthe(\vx_{1:T}, \vz_{1:T}, \bd{f}) = p_\vthe(\bd{f}) p_\vthe(\vz_{1:T}) \prod_{t=1}^T p_\vthe(\vx_t\mid \vz_t, \bd{f}).      
\end{equation}
We discuss the prior distributions $p_\vthe(\bd{f})$ and $p_\vthe(\vz_{1:T})$ separately below.
Each reconstructed frame $\tilde{\vx}_t$, sampled from the frame likelihood $p_\vthe(\vx_t | \bd{f}, \vz_{t})$, 
depends on the corresponding latent variables $\vz_t$ and (optionally) global variables $\bd{f}$. 
We use a Laplace distribution for the frame likelihood, $p_\vthe(\vx_t\mid \vz_t, \bd{f}) = \text{Laplace}\big(\vmu_\vthe(\vz_t, \bd{f}), \lambda^{-1}\boldsymbol{1} \big)$, whose logarithm results in an $\ell_1$ loss which we observe produces sharper images than the $\ell_2$ loss~\citep{DBLP:journals/corr/IsolaZZE16,DBLP:journals/corr/ZhaoGFK15}.
% for autoencoding images~\citep{DBLP:journals/corr/IsolaZZE16,DBLP:journals/corr/ZhaoGFK15}. 

The decoder mean, $\vmu_\vthe(\cdot)$, is a function parametrized by neural networks. Crucially, the decoder is conditioned both on global code $\bd{f}$ and time-local code $\vz_t$. In detail, $(\bd{f}, \vz_{t})$ are combined by a multilayer perceptron (MLP) which is then followed by upsampling transpose convolutional layers to form the mean. More details on the architecture can be found in the appendix. 
After training, the reconstructed frame in image space is obtained from the mean, $\tilde{\vx}_t = \vmu_\vthe(\vz_t, \bd{f})$. 
%The prior distributions $p_\vthe(\bd{f})$ and $p_\vthe(\vz_{1:T})$ will be discussed separately below.

\paragraph{Encoder.} As the inverse of the decoder, the optimal encoder would be the Bayesian posterior $p(\vz_{1:T}, \bd{f}\mid \vx_{1:T})$ of the generative model above, which is analytically intractable. Therefore, we employ amortized variational inference~\citep{blei2017variational,zhang2017advances,marino2018iterative} to predict a distribution over latent codes given the input video,
\begin{equation}
 q_\ff(\vz_{1:T}, \bd{f}\mid \vx_{1:T}) = q_{\ff}(\bd{f}\mid \vx_{1:T})\prod_{t=1}^T q_\ff(\vz_t\mid \vx_t).   
\end{equation}
The global variables $\bd{f}$ are inferred from all video frames in a sequence and may thus contain global information, while $\vz_t$ is only inferred from a single frame $\vx_t$. 

As explained above in Section~\ref{sec:main_lossless}, modifications to standard variational inference are required for further lossless compression into binary. Instead of sampling from Gaussian distributions with learned variances, here we employ fixed-width uniform distributions 
centered at their means: 
$ \tilde{\bd{f}} \sim q_{\ff}(\bd{f}\mid \vx_{1:T})  = \mathcal{U}\big( \hat{\bd{f}} - \frac{1}{2}, \hat{\bd{f}} + \frac{1}{2}  \big)$, $
    \quad
    \tilde{\vz}_t \sim q_\ff(\vz_t\mid \vx_t) = \mathcal{U}\big( \hat{\vz}_t - \frac{1}{2}, \hat{\vz}_t + \frac{1}{2}  \big). 
$
%\begin{align}
%    \tilde{\bd{f}} \sim q_{\ff}(\bd{f}\mid \vx_{1:T})  = \mathcal{U}\big( \hat{\bd{f}} - \frac{1}{2}, \hat{\bd{f}} + \frac{1}{2}  \big) ; \nonumber\\
%    \quad ~~
%    \tilde{\vz}_t \sim q_\ff(\vz_t\mid \vx_t) = \mathcal{U}\big( \hat{\vz}_t - \frac{1}{2}, \hat{\vz}_t + \frac{1}{2}  \big). 
%    \label{eq:post_approx}
%\end{align}

%\begin{figure}[t]
%%\hspace{.5cm}
%\center
%\includegraphics[width=.35\textwidth]{figures/encoder_f.png}
%\caption{\label{fig:encoder_f} Inference network diagram for the global state $\bd{f}$. The features from the video segment are processed by a bi-directional LSTM (with hidden states $\bd{g}^{\bd{f}}$, $\bd{h}^{\bd{f}}$) which is used to infer the global state.}
%\end{figure}

The means are predicted by additional encoder neural networks $\hat{\bd{f}} = \vmu_\ff(\vx_{1:T})$, $\hat{\vz}_t = \vmu_\ff(\vx_t)$ with parameters $\ff$. 
This choice of inference distribution leads exactly to injection of noise with width one centered at the maximally-likely values for the latent variables, described in Section~\ref{sec:main_lossless}. The mean for the global state is parametrized by convolutions over $\vx_{1:T}$, followed by a bi-directional LSTM which is then processed by a MLP. 
%See Fig.~\ref{fig:encoder_f} for the inference diagram. 
The encoder mean for the local state is simpler, consisting of convolutions over each frame followed by a MLP. More details on the decoder architecture is provided in the appendix.

\paragraph{Prior Models.} 
%\CS{could it make sense to merge this into the previous section now?}
The models parametrizing the learned prior distributions are ultimately used as the probability models for entropy coding. The global prior $p_{\vthe}(\bd{f})$ is assumed to be stationary, while $p_{\vthe}(\vz_{1:T})$ consists of a time series model. Each dimension of the latent space has its own density model:
\begin{align}
     p_{\vthe}(\bd{f}) &= \prod_{i}^{\text{dim}(\bd{f})} p_{\vthe} (f^{i} ) * \mathcal{U}(-\frac{1}{2}, \frac{1}{2});
     ~~~~~~~
p_{\vthe}(\vz_{1:T}) = 
     \prod_{t}^T \prod_{i}^{\text{dim}(\vz)} p_{\vthe} (z^{i}_t \mid \vz_{<t}) * \mathcal{U}(-\frac{1}{2}, \frac{1}{2}).\label{eq:priors}
\end{align} 
Above, indices refer to the dimension index of the latent variable. The convolution with uniform noise is to allow the priors to better match the true marginal distribution when working with the box-shaped approximate posterior (see \citet{balle2018variational} Appendix 6.2). This convolution has an analytic form in terms of the cumulative probability density. 

The stationary density $p_{\vthe}(f^{i})$ is adopted from 
\citep{balle2018variational}; it is a flexible non-parametric, fully-factorized model that leads to a good matching between prior and latent code distribution. 
The density is defined by its cumulative and is built out of compositions of nonlinear probability densities, similar to the construction of a normalizing flow \citep{rezende2015variational}.

Two dynamical models are considered to model the sequence $\vz_{1:T}$.
We propose a a recurrent LSTM prior architecture for  $p_{\vthe} (z_t^{i} \mid \vz_{<t})$  
%, 
which conditions on all previous frames in a segment. 
%: $p_{\vthe} (\vz_t^i \mid \bd{c}_t) \equiv p_{\vthe} (z_t^{i} \mid \vz_{<t})$. 
The distribution $p_\vthe$ is taken to be normal with mean and variance predicted by the LSTM. We also considered a simpler model, which we compare against, with a single frame context, $p_{\vthe} (z_t^{i} \mid \vz_{<t}) = p_{\vthe} (z_t^{i} \mid \vz_{t-1})$, which is essentially a deep Kalman filter~\citep{krishnan2015deep}.

\paragraph{Variational Objective.} The encoder (variational model) and decoder (generative model) are learned jointly by maximizing the $\beta$-VAE objective~\citep{higgins2016beta, mandt2016variational}, %which consists of the KL divergence between the approximate and true posterior and takes the form% In compression applications, however, one needs to adjust the trade-off between the bit rate and distortion. This is achieved by the $\beta$-VAE loss~\citep{higgins2016beta, mandt2016variational}, which takes the form (up to constant terms)
\begin{align}
\label{variational:objective}
{\cal L} (\phi,
\theta) = &\mathbb{E}_{\tilde{\bd{f}},\tilde{\vz}_{1:T} \sim q_\phi} [\log p_\vthe(\vx_{1:T}| \tilde{\bd{f}}, \tilde{\vz}_{1:T})] +\beta~\mathbb{E}_{\tilde{\bd{f}},\tilde{\vz}_{1:T} \sim q_\phi} [\log p_\vthe(\tilde{\bd{f}}, \tilde{\vz}_{1:T})].  
%&= \mathbb{E}_{\tilde{\bd{f}},\tilde{\vz}_{1:T} \sim q}  \sum_{t=1}^T  \left\Vert \tilde{\vx}_t - \vx_t\right\Vert_{1} \nonumber \\
%&~~~~~~~~~~~~~~~~~~~~ + \beta H\left[ q_\ff(\tilde{\vz}_{1:T}, \tilde{\bd{f}}\mid \vx_{1:T}), p_\vthe(\tilde{\bd{f}}, \tilde{\vz}_{1:T}) \right].
\end{align}
%where the reconstructed frame $\tilde{\vx}_t = \vmu_\vthe(\tilde{\vz}_t, \tilde{\bd{f}})$. The Laplace parameter $\lambda$ was set to one.

The first term corresponds to the distortion, while second term is the cross entropy between the approximate posterior and the prior. The latter has the interpretation of the expected code length when using the prior distribution $p(\bd{f}, \vz_{1:T})$ to entropy code the latent variables. It is known~\citep{hoffman2016elbo} that this term encourages the prior model to approximate the empirical distribution of codes, $ \mathbb{E}_{\vx_{1:T}} [q(\bd{f}, \vz_{1:T}|\vx_{1:T})]$.
For our choice of generative model, the cross entropy separates into two independent terms $H\big[q_\ff(\bd{f}|\vx_{1:T}), p_\vthe(\bd{f})\big]$ and $H\big[ q_\ff(\vz_{1:T}|\vx_{1:T}), p_\vthe(\vz_{1:T}) \big]$. Note that for our choice of variational distribution, the entropy contribution of $q_\theta$ is constant and is therefore omitted. 
\begin{figure*}[t]
\center
\begingroup
\setlength{\tabcolsep}{6pt} % Default value: 6pt
\renewcommand{\arraystretch}{0.} % Default value: 1
\begin{tabular}{ccc}
 {\small (a) Sprites} & {\small (b) BAIR} & {\small (c) Kinetics} \\
\vspace{-.2cm} 
\hspace{-.2cm} \includegraphics[width=.28\textwidth]{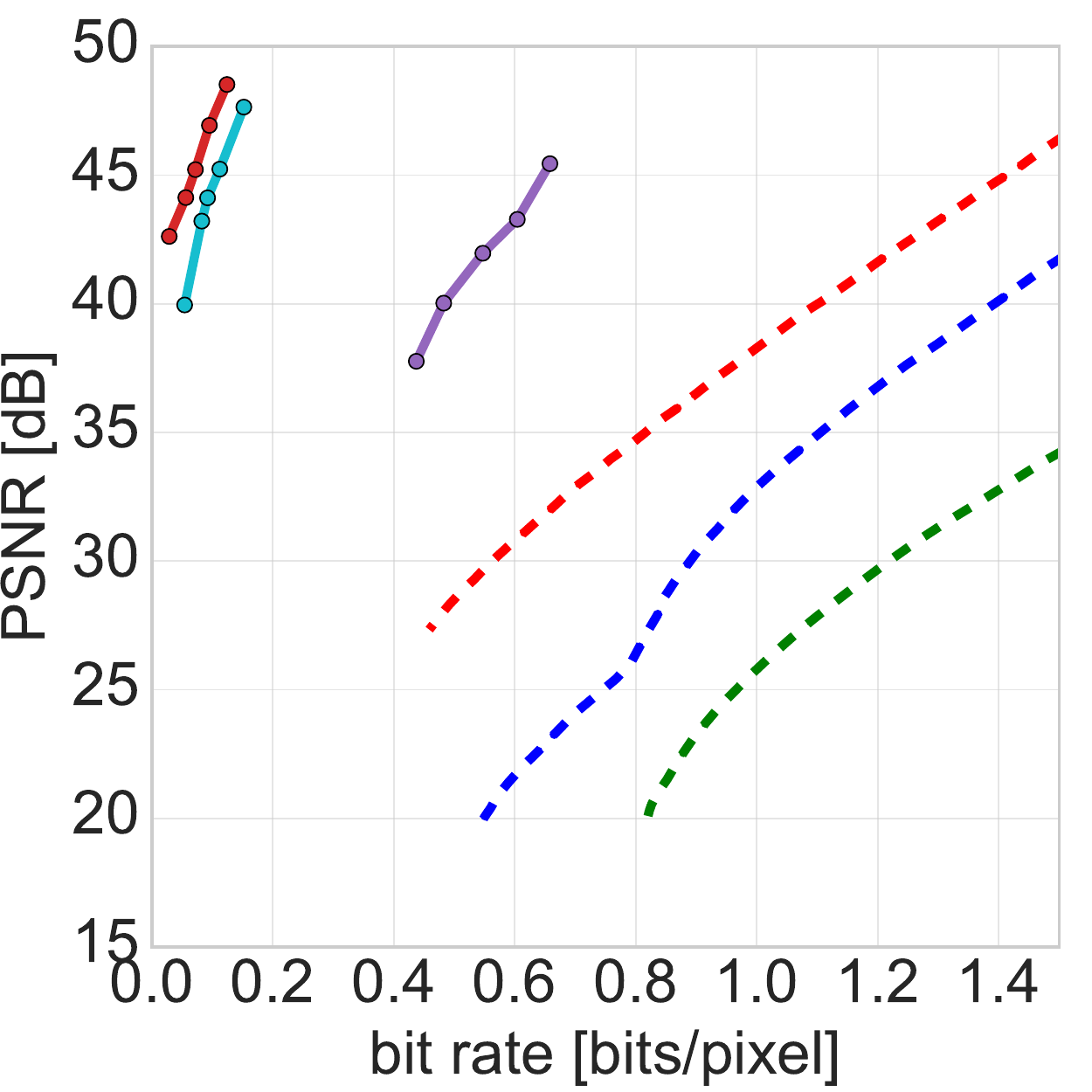}&\hspace{-.2cm}
\includegraphics[width=.28\textwidth]{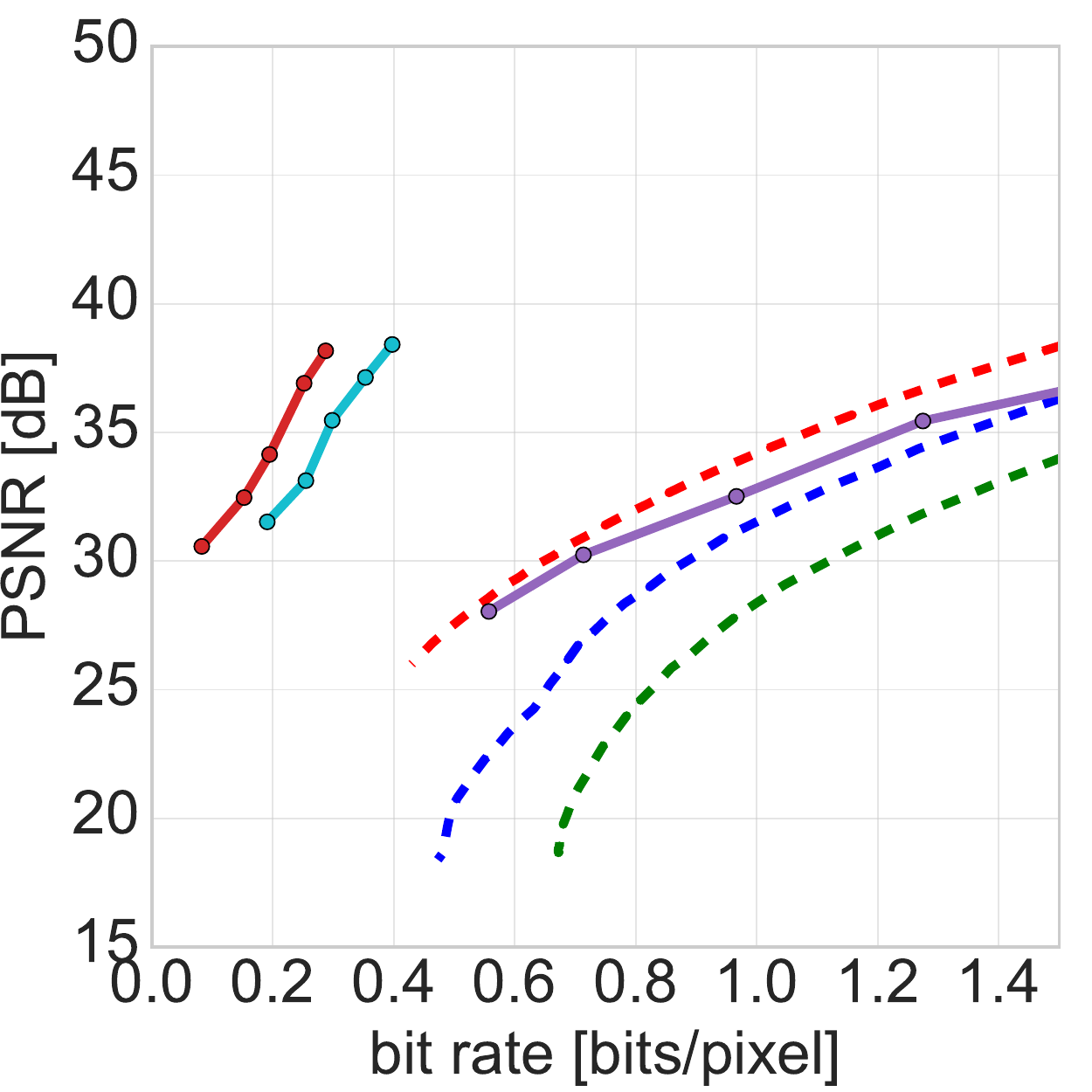}& \hspace{-.2cm}
\includegraphics[width=.28\textwidth]{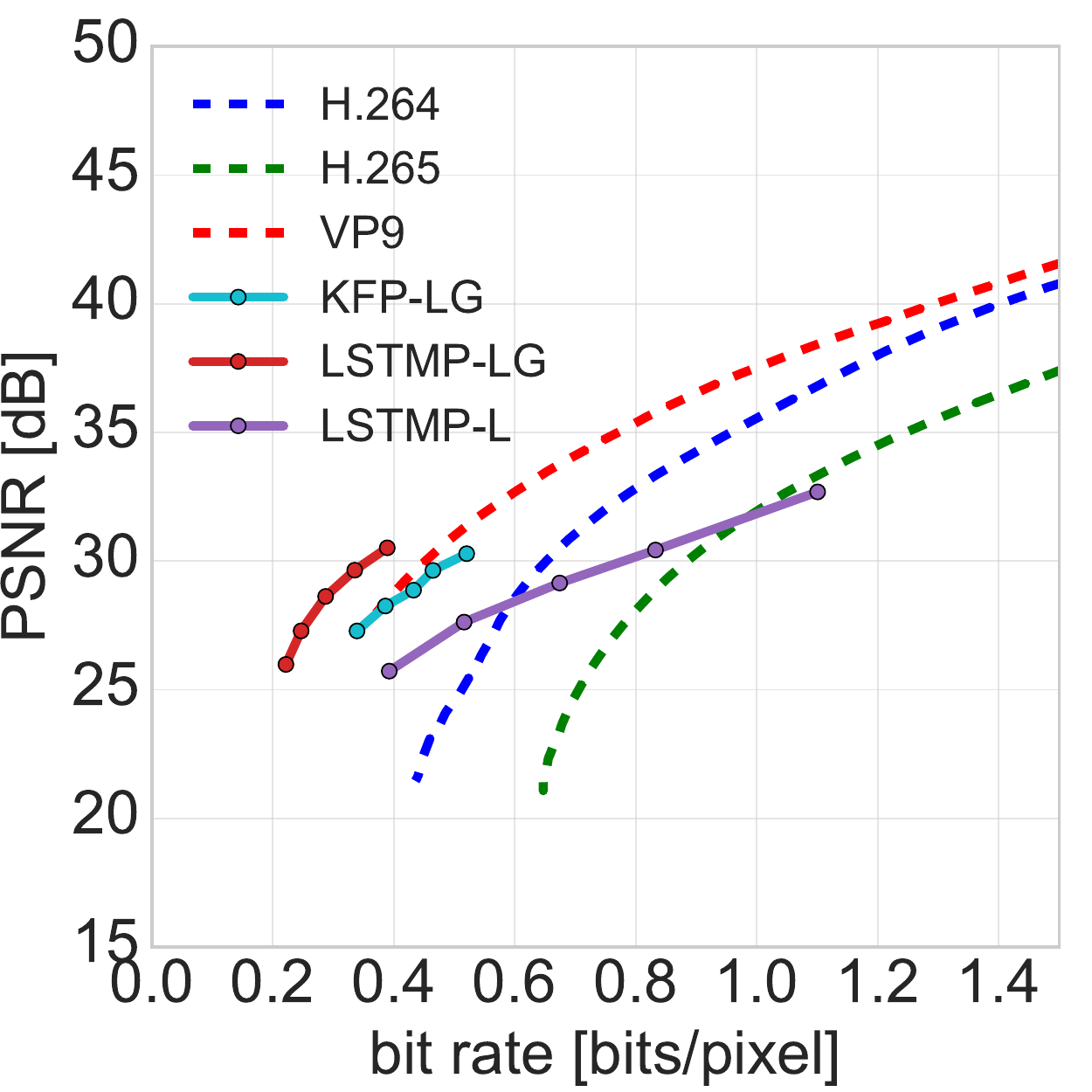}
\end{tabular}
\endgroup
\caption{Rate-distortion curves on three datasets measured in PSNR (higher corresponds to lower distortion). Legend shared. Solid lines correspond to our models, with LSTMP-LG proposed.\label{fig:PSNRcurves}}
\end{figure*}

\section{Experiments}

\label{sec:experiments}
In this section, we present the experimental results of our work. We first describe the datasets, performance metrics, and baseline methods in Section~\ref{sec:datasets}. This is followed by a quantitative analysis in terms of rate-distortion curves in Section~\ref{sec:quantitative} which is followed by qualitative results in Section~\ref{sec:qualitative}.

% We report that our method can achieve extreme compression ratios on videos with specialized content if it is trained on similar videos and that the performance is content-dependent. Our method is also comparable with modern codecs when trained on videos with diverse content. We find that the inclusion of the global state is more efficient than using solely local variables. 
% Implementation details of our proposed models are provided in Appendix \ref{app:architecture}.

\subsection{Datasets, Metrics, and Methods \label{sec:datasets}}

In this work, we train separately on three video datasets of increasing complexity with frame size $64\times64$. \textbf{1) Sprites.} The simplest dataset consists of videos of Sprites characters from an open-source video game project, which is used in \citep{reed2015deep,mathieu2016disentangling, li2018deep}. The videos are generated from a script that samples the character action, skin color, clothing, and eyes from a collection of choices and have an inherently low-dimensional description (\textit{i.e.} the script that generated it). \textbf{2) BAIR.} BAIR robot pushing dataset \citep{ebert2017self} consists of a robot pushing objects on a table, which is also used in \citep{babaeizadeh2017stochastic,denton2018stochastic,lee2018stochastic}. The video is more realistic and less sparse, but the content is specialized since all scenes contain the same background and robot, and the depicted action is simple since the motion is described by a limited set of commands sent to the robot. The first two datasets are uncompressed and no preprocessing is performed. \textbf{3) Kinetics600.} The last dataset is the Kinetics600 dataset \citep{kay2017kinetics} which is a diverse set of YouTube videos depicting human actions. The dataset is cropped and downsampled, which removes compression artifacts, to $64 \times 64$.

\textbf{Metrics.} Evaluation is based on bit rate in bits per pixel (bpp) and distortion measured in average frame peak signal-to-noise ratio (PSNR), which is related to the frame mean square error. In the appendix, we also report on multi-scale structural similarity (MS-SSIM) \citep{wang2004image} which is a perception-based metric that approximates the change in structural information. 

%\begin{figure*}[t]
%\begin{center}
%\begingroup
%\setlength{\tabcolsep}{0pt} % Default value: 6pt
%\renewcommand{\arraystretch}{0.} % Default value: 1
%\begin{tabular}{ccc}
%\includegraphics[width=.33\textwidth]{figures/sprites-entropy.pdf} &
%\includegraphics[width=.33\textwidth]{figures/bair-entropy.pdf} &
%\includegraphics[width=.33\textwidth]{figures/kinetics-entropy.pdf} \\
%  {\small (a) Sprites} & {\small (b) BAIR } & {\small (c) Kinetics}  \\
%  \end{tabular}
%\endgroup
%\end{center}
%\caption{Average bits of information stored in $\vf$ and $\vz_{1:T}$ with PSNR 43.2, 37.1, 30.3 dB for different models in (a, b, c). Entropy drops with the frame index as the models adapt to the video sequence. \label{fig:entropy}}
% \end{figure*}

\textbf{Comparisons.}
We wish to study the performance of our proposed local-global architecture with LSTM prior (LSTMP-LG) by comparing to other approaches. To study the effectiveness of the global state, we introduce our baseline model LSTMP-L which has only local states with LSTM prior $p_{\vthe} (\vz_t \mid \vz_{<t})$. To study the efficiency of the predictive model, we show our baseline model KFP-LG which has both global and local states but with a weak predictive model $p_{\vthe} (\vz_t \mid \vz_{t-1})$, a deep Kalman filter~\citep{krishnan2015deep}. 
We also provide the performance of H.264, H.265, and VP9 codecs. Traditional codecs are not optimized for low-resolution videos. However, their performance is far superior to neural or classical image compression methods (applied to compress video frame by frame), so their performance is presented for comparison. Codec performance is evaluated using the open source FFMPEG implementation in constant rate mode and distortion is varied by adjusting the constant rate factor. Unless otherwise stated, performance is tested on videos with 4:4:4 chroma sampling and on test videos with $T=10$ frames. Comparisons with classical codec performance on longer videos is shown in the appendix.

% \citep{wiegand2003overview,sullivan2012overview,mukherjee2015technical}.

\begin{figure*}[t]
\begin{center}
\begingroup
\setlength{\tabcolsep}{0pt} % Default value: 6pt
\renewcommand{\arraystretch}{0.} % Default value: 1
\begin{tabular}{ccc}
\includegraphics[width=.33\textwidth]{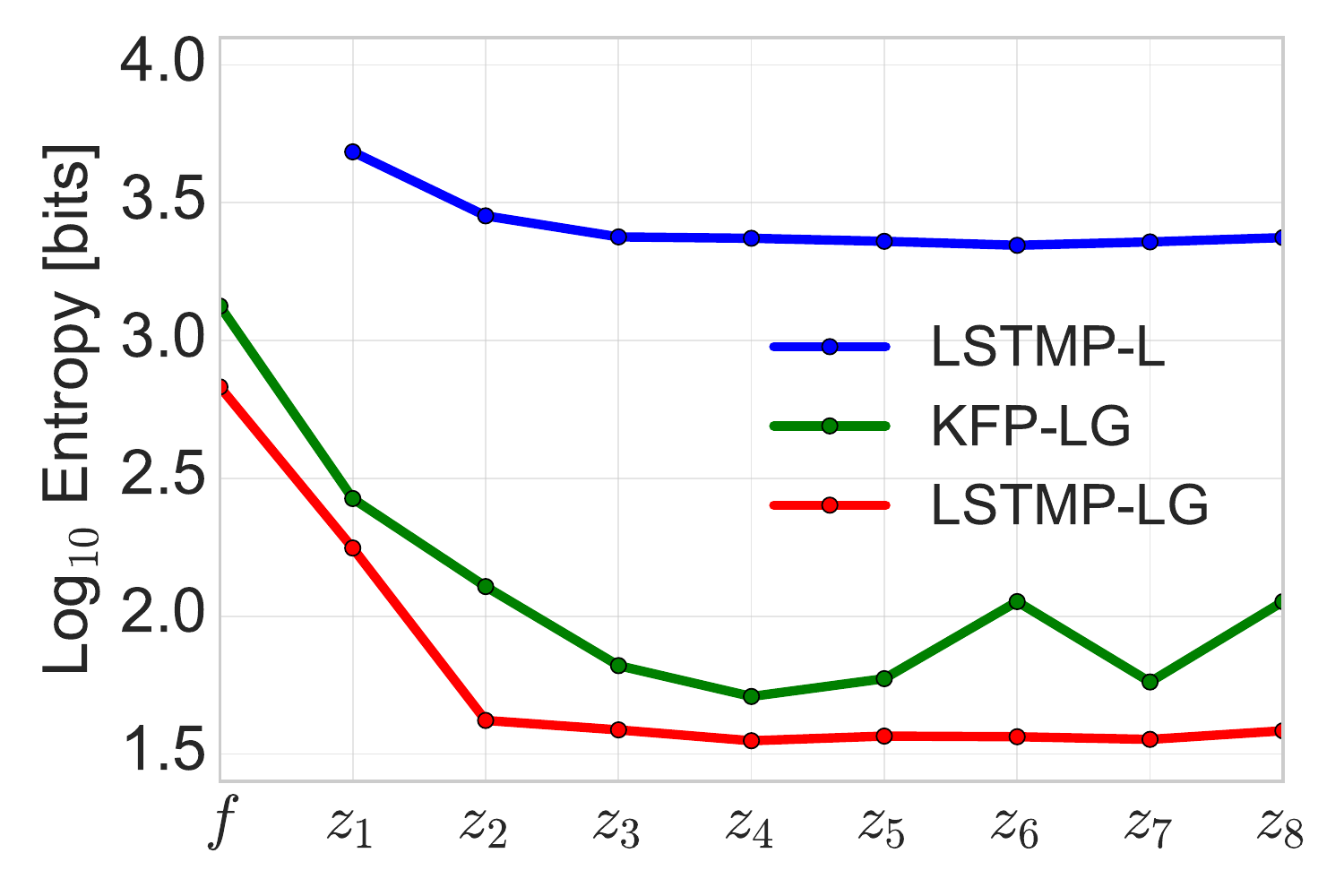} &
\includegraphics[width=.33\textwidth]{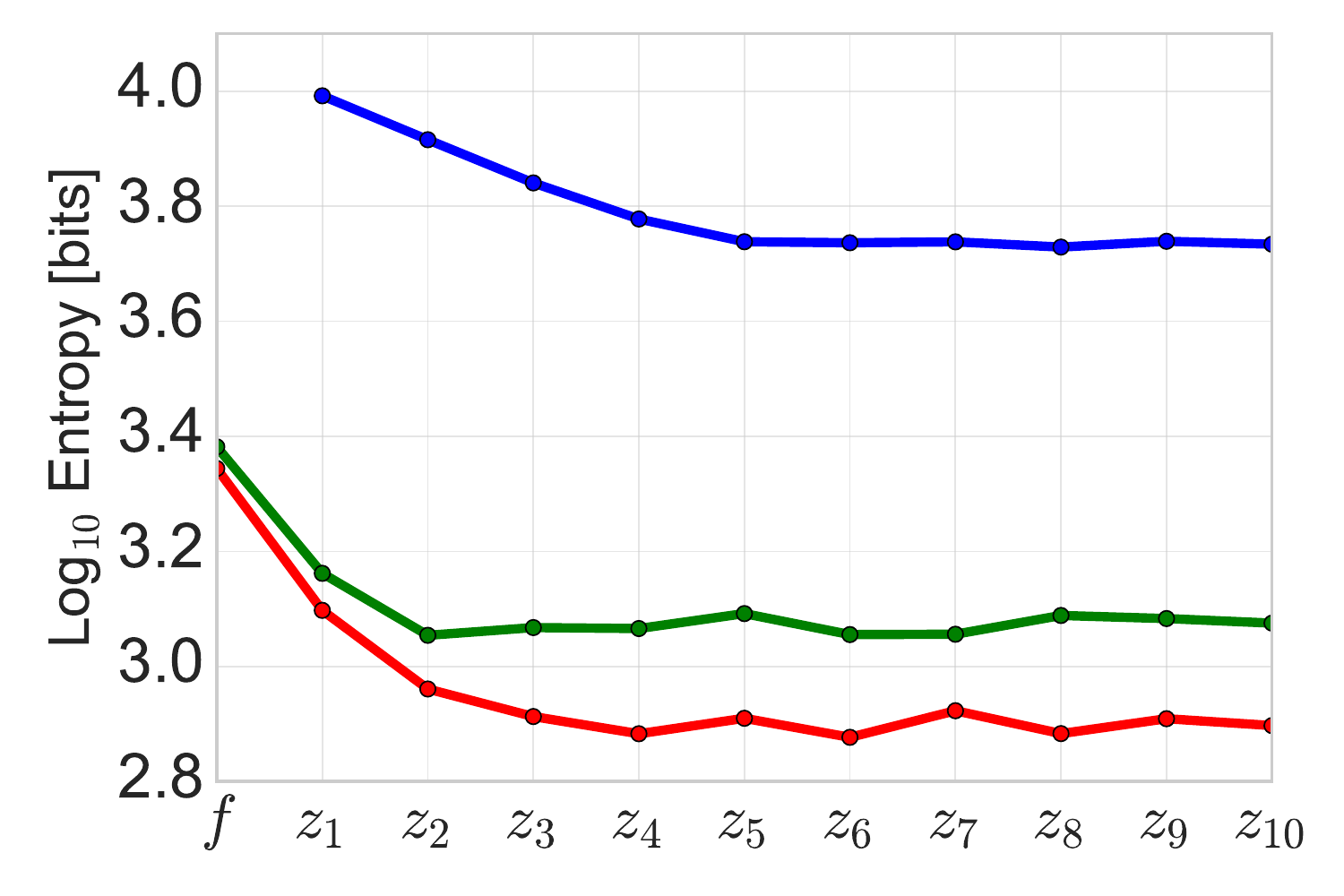} &
\includegraphics[width=.33\textwidth]{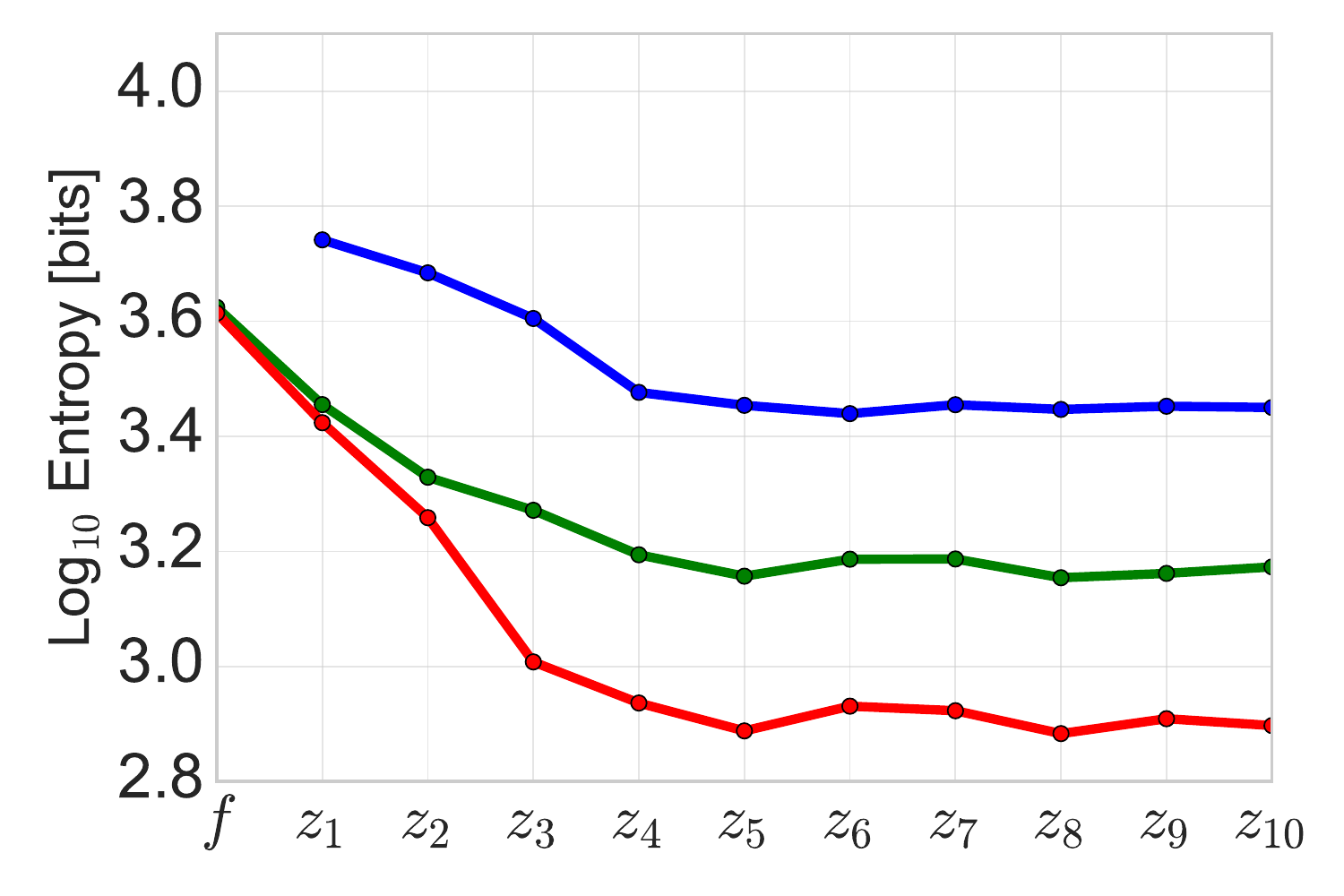} \\
  {\small (a) Sprites} & {\small (b) BAIR } & {\small (c) Kinetics} 
  \end{tabular}
\endgroup
\end{center}
\vspace{-.3cm}
\caption{Average bits of information stored in $\vf$ and $\vz_{1:T}$ with PSNR 43.2, 37.1, 30.3 dB for different models in (a, b, c). Entropy drops with the frame index as the models adapt to the  sequence. \label{fig:entropy}}
 \end{figure*}

\subsection{Quantitative Analysis: Rate-Distortion Tradeoff \label{sec:quantitative}}
Quantitative  performance is evaluated in terms of rate-distortion curves.
%Quantitative compression performance is evaluated in terms of rate-distortion theory. This characterizes the trade-off between the binary representation size and average distortion. 
For a fixed quality setting, a video codec produces an average bit rate on a given dataset. By varying the quality setting, a curve is traced out in the rate-distortion plane. Our curves are generated by varying $\beta$ (Eq.~\ref{variational:objective}).

The rate-distortion curves for our method, trained on three datasets and measured in PSNR, are shown in Fig.~\ref{fig:PSNRcurves}. Higher curves indicate better performance. From the Sprites and BAIR results, one sees that our method has the ability to  dramatically  outperform traditional codecs when focusing on specialized content. By training on videos with a fixed content, the model is able to learn an efficient representation for such content, and the learned priors capture the empirical data distribution well. The results from training on the more diverse Kinetics videos also outperform or are competitive with standard codecs and better demonstrate the performance of our method on general content videos. Similar results are obtained with respect to MS-SSIM (appendix). 

% We also plot the average frame MS-SSIM with respect to the bit rate to quantitatively compare our models with traditional codecs. From Fig.~\ref{fig:msssim}, we can see that our LSTMP-LG method saves significantly more bits when trained on specialized content dataset (Sprites and BAIR) and achieves competitive result with respect to MS-SSIM when trained on general content dataset. 

The first observation is that the LSTM prior outperforms the deep Kalman filter prior  in all cases. This is because the LSTM model has more context, allowing the predictive model to be more certain about the trajectory of the local latent variables, which in turn results in shorter code lengths. 
We also observe that the local-global architecture (LSTMP-LG) outperforms the local architecture (LSTMP-L) on all datasets.
%, demonstrating the usefulness of a hybrid approach which partially encodes the entire video segment in a global state along with extra frame-by-frame information stored as a sequence. 
The VAE encoder has the option to store information in local or global variables. The local variables are modeled by a temporal prior and can be efficiently stored in binary if the sequence $\vz_{1:T}$ can be sequentially predicted from the context. The global variables, on the other hand, provide an architectural approach to removing temporal redundancy since the entire segment is stored in one global state without temporal structure. 
 
 During training, the VAE learns to utilize the global and local information in the optimal way. The utilization of each variable can be visualized by plotting the average code length of each latent state, which is shown in Fig.~\ref{fig:entropy}. The VAE learns to significantly utilize the global variables even though $\text{dim}(\vz)$ is sufficiently large to store the entire content of each individual frame. This provides further evidence that it is more efficient to incorporate global inference over several frames. The entropy in the local variables initially tends to decrease as a function of time, which supports the benefits from our predictive models. Note that our approach relies on sequential decoding, prohibiting a bi-directional LSTM prior model for the local state. 

\subsection{Qualitative Results \label{sec:qualitative}}
\begin{figure*}[t]
\begin{center}
\begingroup
\setlength{\tabcolsep}{.1pt} % Default value: 6pt
\begin{tabular}{ccc}
 \multicolumn{3}{c}{{\small Ours (\textbf{38.1} dB @ \textbf{0.29} bpp)}} \\
% {\small Ours} & {\small H.265} & {\small VP9}\\
\includegraphics[height=.15\textwidth]{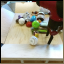}&
\includegraphics[height=.15\textwidth]{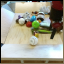} &
\includegraphics[height=.15\textwidth]{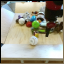} \\
 \multicolumn{3}{c}{{\small VP9 (\textbf{25.7} dB @ \textbf{0.44} bpp)}} \\
\includegraphics[height=.15\textwidth]{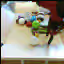}&
\includegraphics[height=.15\textwidth]{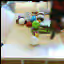} &
\includegraphics[height=.15\textwidth]{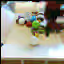} \\
{\small t=1} & {\small t=5} & {\small t=10} \\
\end{tabular} 
~~~~~\begin{tabular}{ccc}
{\small Original} & {\small Ours (\textbf{0.39} bpp)} & {\small VP9 (\textbf{0.39} bpp)} \\
% {\small Ours} & {\small H.265} & {\small VP9}\\
\includegraphics[height=.15\textwidth]{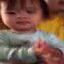}&
\includegraphics[height=.15\textwidth]{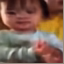} &
\includegraphics[height=.15\textwidth]{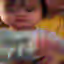} \\
& {\small \textbf{32.0} dB }& {\small \textbf{29.3} dB} \\
\includegraphics[height=.15\textwidth]{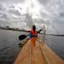}&
\includegraphics[height=.15\textwidth]{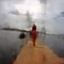} &
\includegraphics[height=.15\textwidth]{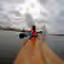} \\
& {\small \textbf{30.1} dB }& {\small \textbf{30.8} dB} \\
% {\small t=1} & {\small t=14} & {\small t=30} \\
\end{tabular} 
\endgroup
\vspace{-.3cm}
 \caption{Compressed videos by our LSTMP-LG model and VP9 in the low bit rate regime (measured in bpp). Our approach achieves better quality (measured in dB) on specialized content (BAIR, left) and comparable visual quality on generic video content (Kinetics, right) compared to VP9. \label{fig:combined}
}  
\end{center}
\end{figure*}

We have shown that a deep neural approach (LSTMP-LG architecture) can achieve competitive results with traditional codecs with respect to PSNR or MS-SSIM (see appendix) metrics overall on low-resolution videos. Test videos from the Sprites and BAIR datasets after compression with our method are shown in Fig.~\ref{fig:sprite} and Fig.~\ref{fig:combined} (left), respectively, and compared to modern codec performance. Our method achieves a superior image quality at a significantly lower bit rate than H.264/H.265 and VP9 on these specialized content datasets. This is perhaps expected since traditional codecs cannot learn efficient representations for specialized content. 
Furthermore, fine-grained motion is not accurately predicted with block motion estimation. 
The artifacts from our method are displayed in Fig.~\ref{fig:combined} (right). Our method tends to produce blurry video in the low bit-rate regime but does not suffer from the block artifacts present in the H.265/VP9 compressed video. 

\section{Conclusions}
\label{sec:conclusions}
We have proposed a deep generative modeling approach to video compression. Our method simultaneously learns to transform the original video into a lower-dimensional representation as well as the temporally-conditioned probabilistic model for entropy coding. 
The best performing proposed architecture splits up the latent code into global and local variables and yields competitive results on low-resolution videos. For video sources with specialized content, deep generative video coding allows for a significant increase in coding performance, as our experiment on BAIR suggests. This could be interesting for transmitting specialized content such as teleconferencing. 

Our experimental analysis focused on small-scale videos. One future avenue is to design alternative priors that better scale to full-resolution videos, where the dimension of the latent representation must scale with the resolution of the video in order to achieve high quality reconstruction. For the local/global architecture that we investigated experimentally, the GPU memory limits the maximum size of the latent dimension due to the presence of fully-connected layers to infer global and local states. While being efficient for small videos in the strongly compressed regime, this effectively limits the maximum achievable image quality. Future architectures may focus more on fully convolutional components. Besides a different temporal prior, the proposed coding scheme will remain the same.

Since our approach uses a learned prior for entropy coding, this suggests that improved compression performance can be achieved by improving video prediction.  In future work, it will be interesting to see how our model will work with more efficient predictive models for full-resolution videos. It is also interesting to think about comparisons between deterministic and stochastic approaches to neural compression. We argue that by modeling the full data distribution of each frame, a probabilistic approach should be able to achieve shorter code lengths for fat-tailed and skewed data distributions than maximum-likelihood based compression methods. 
Furthermore, there is much potential in investigating how additional `side information' could aid the predictive model. Thus we think that our work is a first step into a new direction for video coding which opens up several exciting avenues for future work. 

This work was completed while Jun Han and Stephan Mandt were affiliated with Disney Research.

%\section*{References}

%References follow the acknowledgments. Use unnumbered first-level heading for
%the references. Any choice of citation style is acceptable as long as you are
%consistent. It is permissible to reduce the font size to \verb+small+ (9 point)
%when listing the references. {\bf Remember that you can use more than eight
%  pages as long as the additional pages contain \emph{only} cited references.}
%\medskip
%
%\small
%
%[1] Alexander, J.A.\ \& Mozer, M.C.\ (1995) Template-based algorithms for
%connectionist rule extraction. In G.\ Tesauro, D.S.\ Touretzky and T.K.\ Leen
%(eds.), {\it Advances in Neural Information Processing Systems 7},
%pp.\ 609--616. Cambridge, MA: MIT Press.
%
%[2] Bower, J.M.\ \& Beeman, D.\ (1995) {\it The Book of GENESIS: Exploring
%  Realistic Neural Models with the GEneral NEural SImulation System.}  New York:
%TELOS/Springer--Verlag.
%
%[3] Hasselmo, M.E., Schnell, E.\ \& Barkai, E.\ (1995) Dynamics of learning and
%recall at excitatory recurrent synapses and cholinergic modulation in rat
%hippocampal region CA3. {\it Journal of Neuroscience} {\bf 15}(7):5249-5262.

\bibliography{iclr2019_conference}
\bibliographystyle{iclr2019_conference}

\newpage
\appendix

\section{Model architecture \label{app:architecture}}

The specific implementation details of our model are now described. We describe the two baseline models, LSTMP-L and KFP-LG, and the best-performing LSTMP-LG model. 

\textbf{LSTMP-L.} Our proposed baseline model LSTMP-L contains only local latent variables $\bd{z}_t$ (the global state $\bd{f}$ is omitted). This model is introduced to study the efficiency of the global state for removing temporal redundancy. The local state for each frame $\bd{z}_t$ is inferred from each frame $\bd{x}_t$. LSTMP-L employs similar encoder and decoder architectures as \citet{balle2016end}. The encoder $\vmu_\ff (\vx_t)$ infers each $\vz_t$ independently by a five-layer convolutional network. For layer $\ell=1$, the stride is 4, while a stride of 2 is used for layer $\ell=2, 3, 4, 5$. The padding is 1 and the kernel size is $4\times 4$ for all layers. The number of filters used for the Sprites video, for $\ell=1, 2, 3, 4, 5$, are 192, 256, 512, 512 and 1024, respectively. For the more realistic video (BAIR and Kinetics video), the number of filters used at layer $\ell=1, 2, 3, 4, 5$ are 192, 256, 512, 1024 and 2048, respectively. The decoder $\vmu_\vthe(\vz_t)$ is symmetrical to the encoder $\vmu_\ff (\vx_t)$. With this architecture, the dimension of the latent state $\vz_t$ is 1024 for Sprites and 2048 for BAIR and Kinetics video. The prior for the latent state corresponding to the first frame, $p_\vthe(\vz_1)$, is parametrized by the same density model defined on Appendix 6.1 of \citet{balle2018variational}. The conditional prior $p_{\vthe} (\vz_t \mid \vz_{<t})$ is parameterized by a normal distribution convolved with uniform noise. The means and (diagonal) covariance of the normal distribution are predicted by an LSTM with hidden state dimension equal to the dimension of the latent state $\vz_t$.

\textbf{LSTMP-LG.} LSTMP-LG is our proposed model in this paper which uses an efficient latent representation by splitting latent states into both global states and local states as well as the use of an effective LSTM predictive model for entropy coding. Now we describe the inference network. The two encoders $\vmu_\ff(\vx_{1:T})$ and $\vmu_\ff(\vx_t)$ begin with a convolutional architecture to extract feature information. The global state $\bd{f}$ is inferred from
all frames by processing the output of the convolutional layers over $\vx_{1:T}$ with a bi-directional LSTM architecture (note this LSTM is used for inference not entropy coding). This allows $\bd{f}$ to depend on features from the entire segment. For the local state, the individual frame $\vx_t$ is passed through the convolutional layers of $\vmu_\ff (\vx_t)$ and a two-layer MLP infers $\vz_t$ from the feature information of the individual frame. The decoder $\vmu_\vthe(\vz_t, \bd{f})$ first combines $(\vz_t, \bd{f})$ with a multilayer perceptron (MLP) and then upsamples with a deconvolutional network. The prior models $p_\vthe(\vf)$ and $p_\vthe(\vz_1)$ are parametrized by the density model defined in Appendix 6.1 of \citet{balle2018variational}. The conditional prior $p_{\vthe} (\vz_t \mid \vz_{<t})$ in the LSTMP-LG architecture is modeled by a normal distribution which is convolved with uniform noise. The means and covariance of the normal distribution are predicted by an additional LSTM.

Both encoders $\vmu_\ff (\cdot)$ have 5 convolutional (downsampling) layers. For layer $\ell=1, 2, 3, 4$, the stride and padding are 2 and 1, respectively, and the convolutional kernel size is 4$\times$4. The number of channels for layer $\ell=1, 2, 3, 4$ are 192, 256, 512, 1024. Layer 5 has kernel size 4, stride 1, padding 0, and 3072 channels. The decoder architecture $\vmu_\vthe$ is chosen to be asymmetric to the encoder with convolutional layers replaced with deconvolutional (upsampling) layers. For the Sprites toy video, the dimensions of $\boldsymbol{z}$, $\boldsymbol{f}$, and hidden state $\boldsymbol{h}$ are 64, 512 and 1024, respectively. For less sparse videos (BAIR and Kinetics600), the dimensions of  $\boldsymbol{z}$, $\boldsymbol{f}$, and LSTM hidden state $\boldsymbol{h}$ are 256, 2048 and 3072, respectively. 

\begin{figure*}[t]
\begin{center}
\begingroup
\setlength{\tabcolsep}{0pt} % Default value: 6pt
\renewcommand{\arraystretch}{0.} % Default value: 1
\begin{tabular}{cccc}
\includegraphics[height=.23\textwidth]{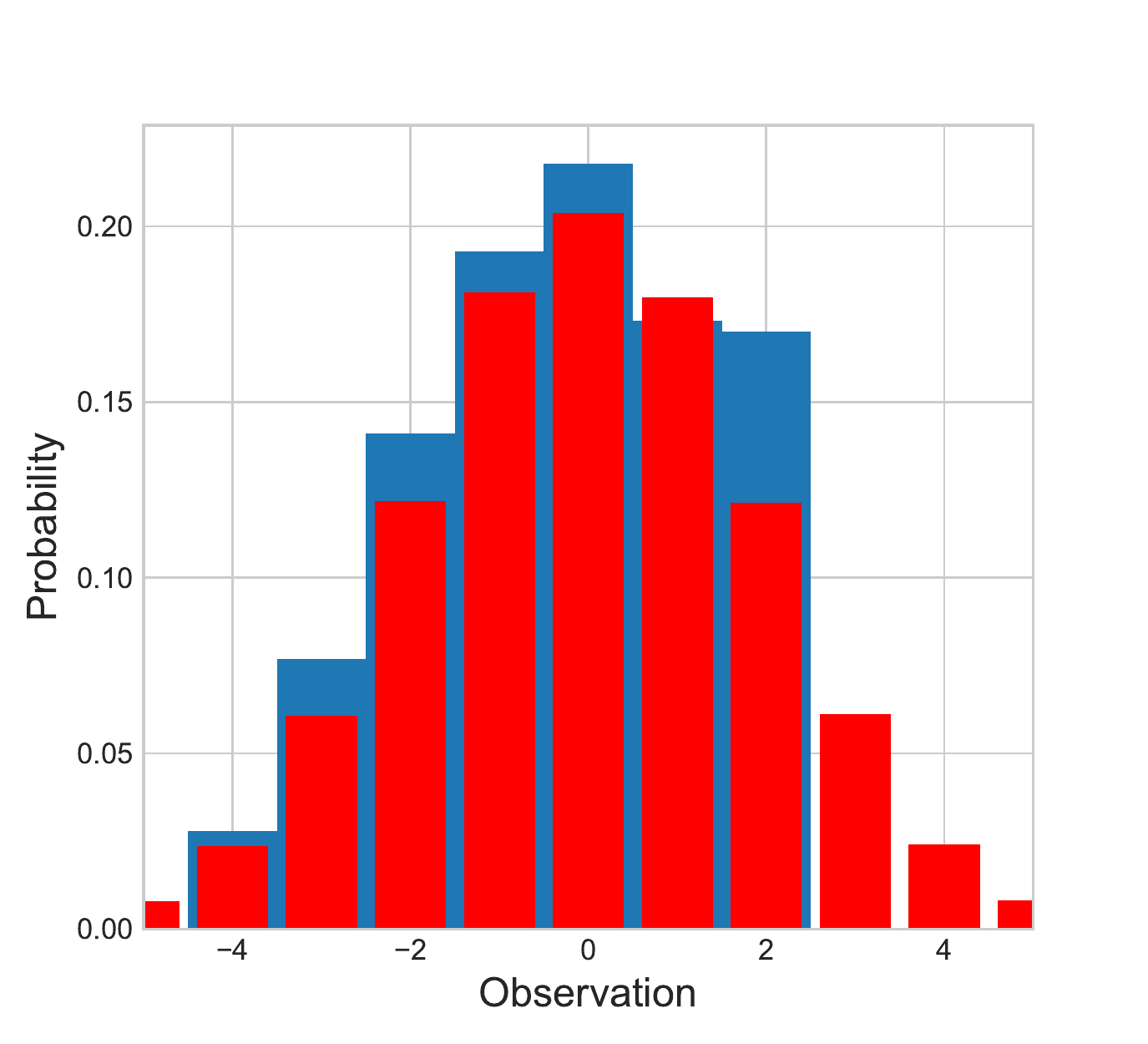}&
\includegraphics[height=.23\textwidth]{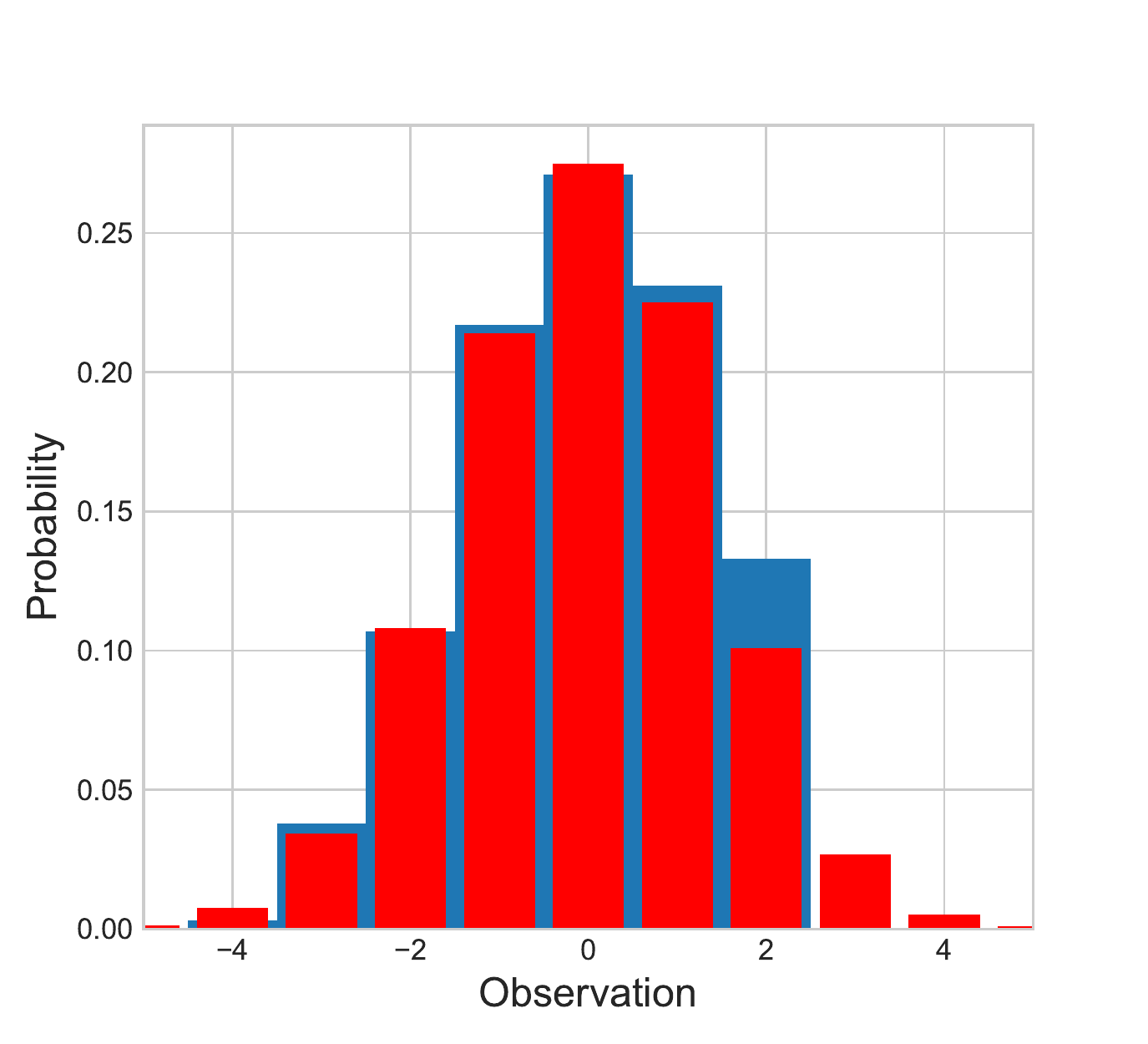}&
 \includegraphics[height=.23\textwidth]{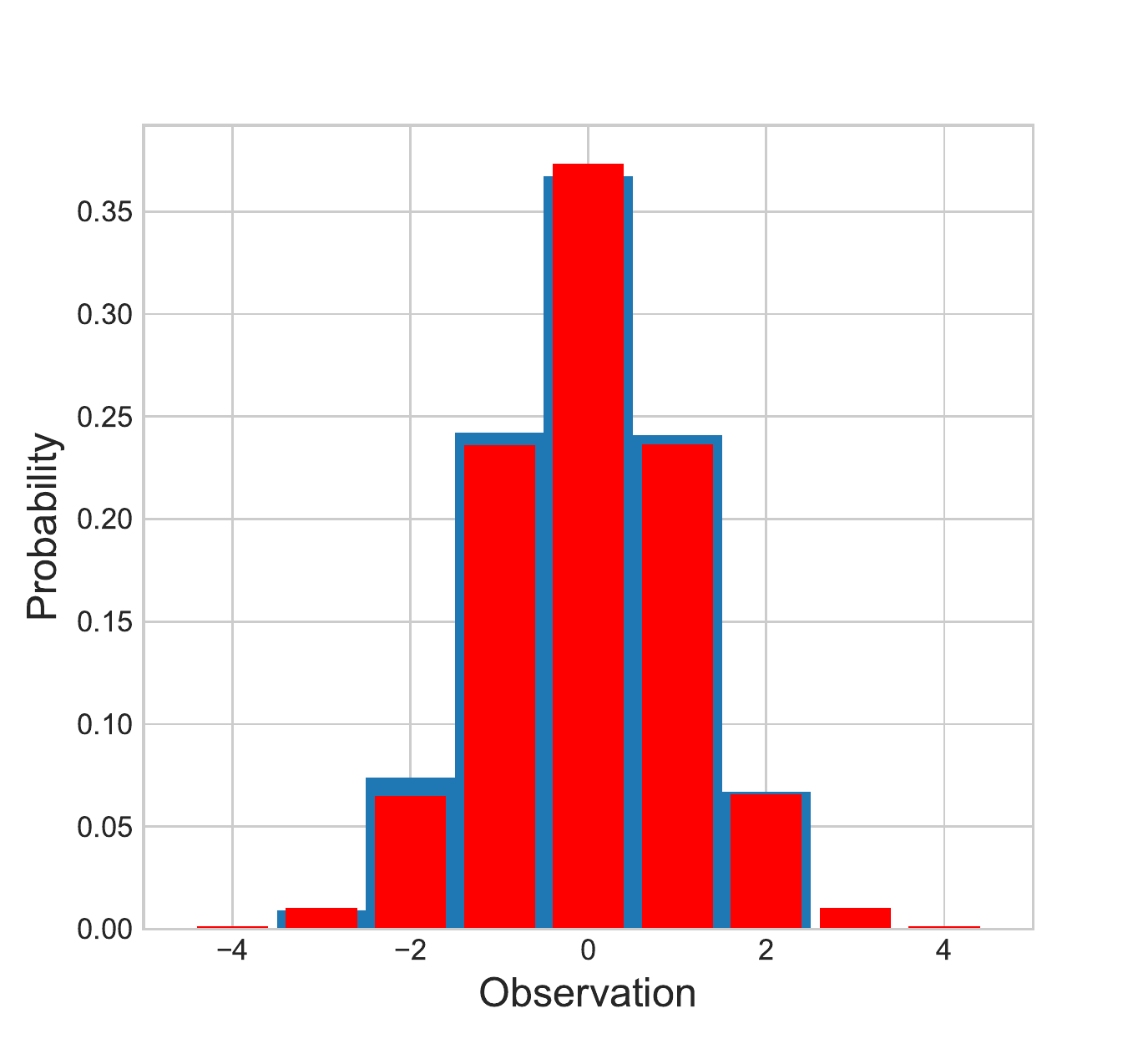}& 
\includegraphics[height=.23\textwidth]{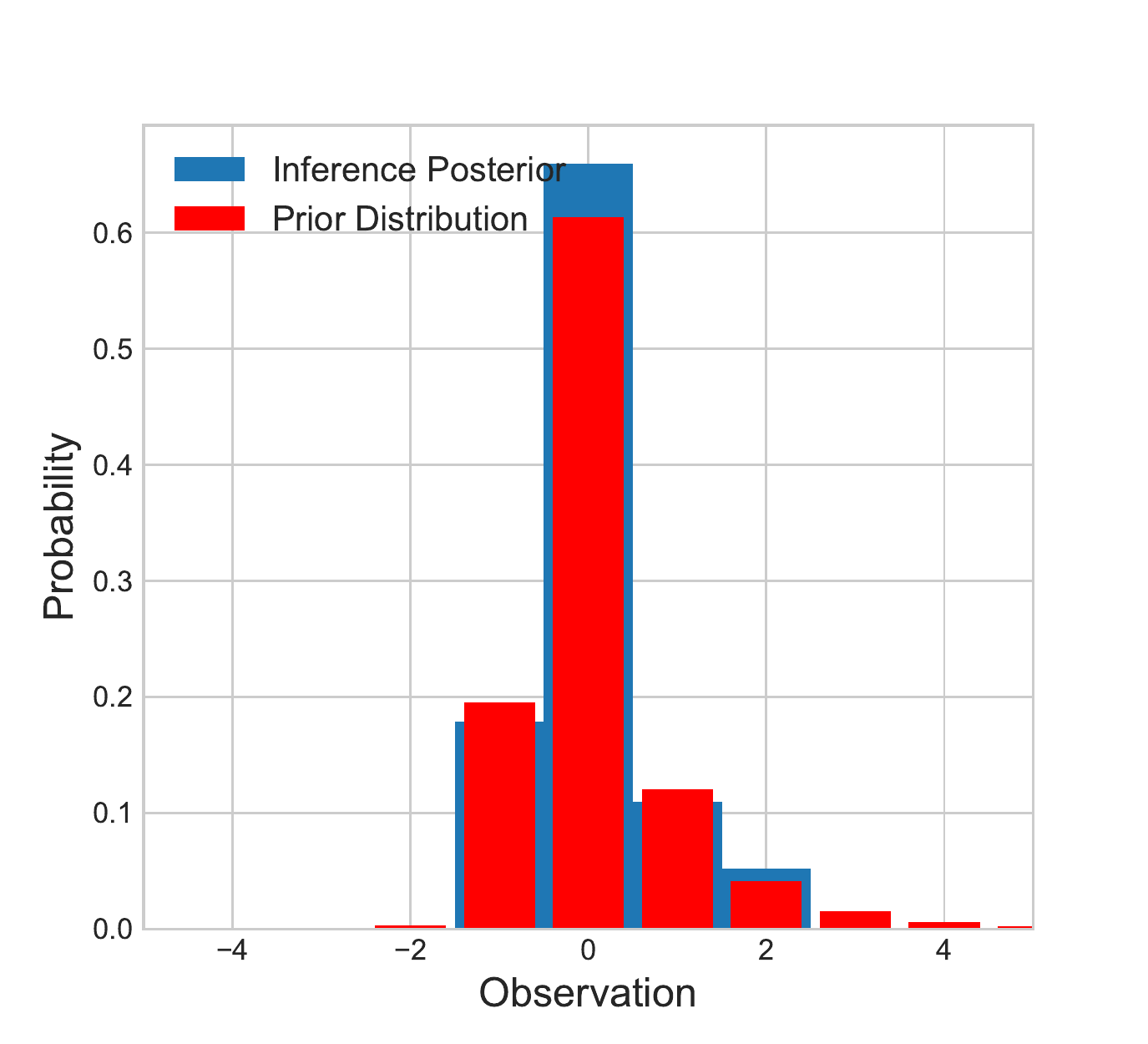}\\
{\small (a) $p_\vthe(f^i)$} & {\small (b) $p_\vthe(f^i)$} &{\small (c) $p_\vthe(z_1^i)$} & {\small (d) $p_\vthe(z_1^i)$}
  \end{tabular}
\endgroup
\end{center}
\caption{\label{fig:fprior} Empirical distributions of the posterior of inference model and ground truth prior model in one specific rate-distortion BAIR example.}
 \end{figure*}

\textbf{KFP-LG.} 
KFP-LG is also a proposed baseline model which incorporates both the global state $\bd{f}$ and local latent $\bd{z}_t$ but uses a weaker deep Kalman filter predictive model $p_{\vthe} (\vz_t \mid \vz_{t-1})$ for entropy coding. The main purpose of the KFP-LG model is to compare to the LSTMP-LG model which has a longer memory. The conditional prior $p_{\vthe} (\vz_t \mid \vz_{t-1})$ in KFP-LG is described by a normal distribution with mean and variance that are parametrized by a three-layer MLP. The dimension at each layer of MLP is the same as the dimension of the latent state $\vz_t$. KFP-LG has the same encoder and decoder structures as the proposed LSTMP-LG model aforementioned. The only difference between KFP-LG and LSTMP-LG is that they employ different prior models for conditional entropy coding.

\section{Latent variable distribution visualization \label{latent:visual}}

In this section, we visualize the distribution of our prior model and compare to the empirical distribution of the posterior of the inference model estimated from data. In Fig.~\ref{fig:fprior}, we show the learned priors and the empirically observed posterior over two dimensions of the global latent state $\vf$ and local latent state $\vz$ in order to demonstrate that the prior is capturing the correct empirical distribution. From Fig.~\ref{fig:fprior}, we can see that the learned priors $p_\vthe(\bd{f})$ and $p_\vthe(\bd{z}_1)$ match the empirical data distributions well, which leads to low bit rate encoding of the latent variables. As the conditional probability model $p_{\vthe} (\vz_t \mid \vz_{<t})$ is high-dimensional, we do not display this distribution.

% \section{MS-SSIM RATE-DISTORTION PERFORMANCE \label{app:msssim}}
% \label{app:more-results}
\section{Additional performance evaluation \label{app:msssim}}
\label{app:more-results}
\begin{figure*}[h]
\begingroup
\setlength{\tabcolsep}{0pt} % Default value: 6pt
\renewcommand{\arraystretch}{0.} % Default value: 1
\begin{tabular}{ccc}
 {\small (a) Sprites} & {\small (b) BAIR} & {\small (c) Kinetics} \\
\includegraphics[width=.33\textwidth]{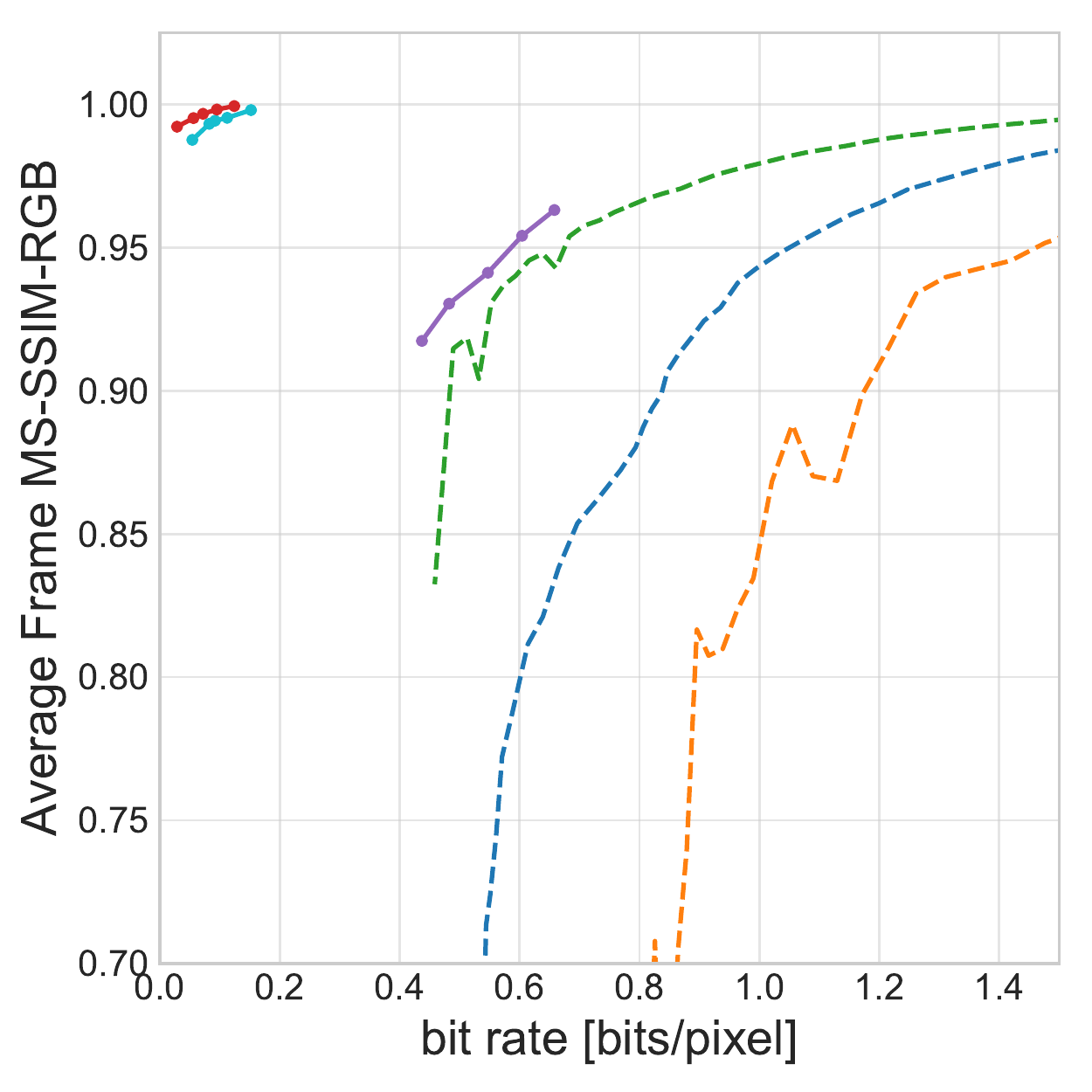}&
\includegraphics[width=.33\textwidth]{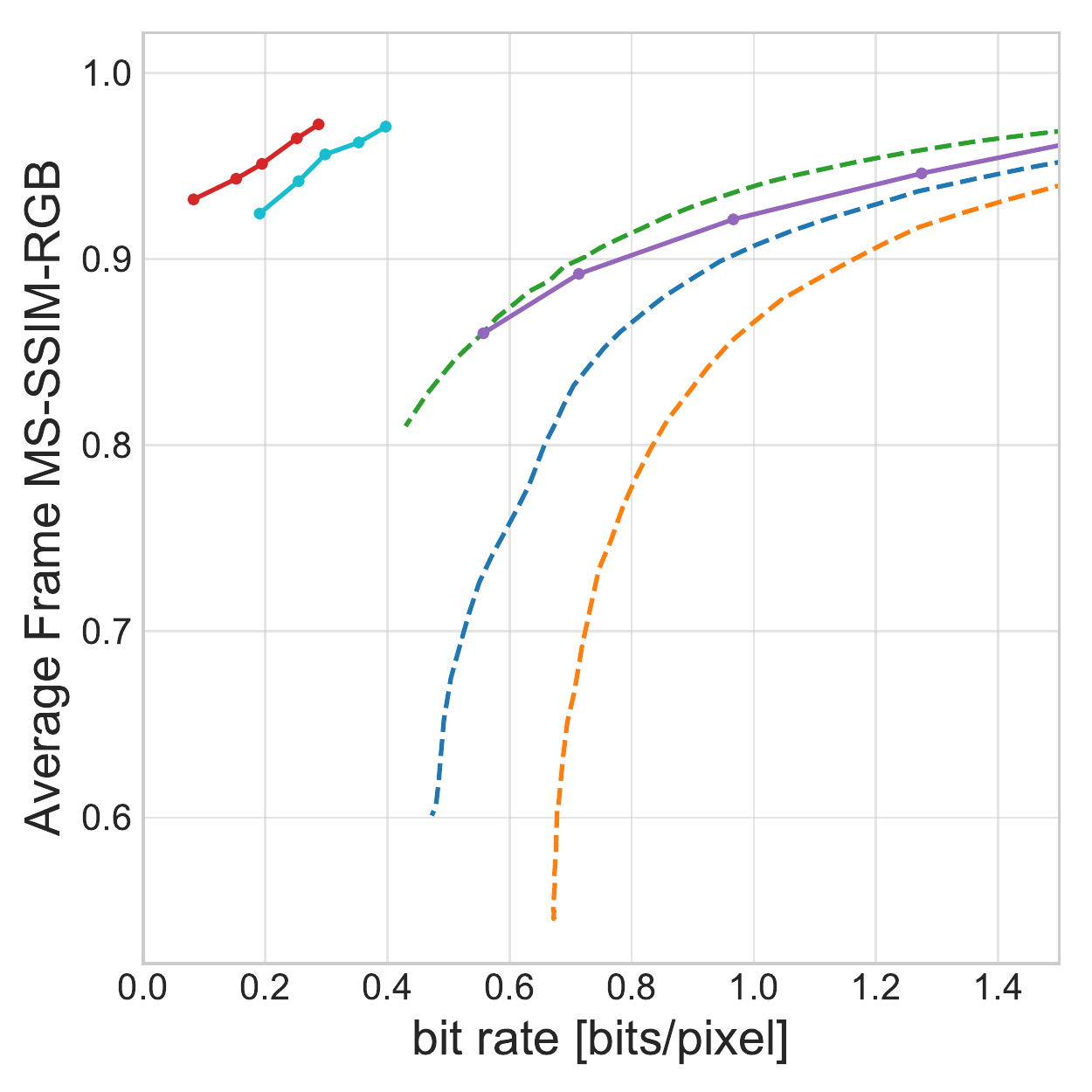}&
\includegraphics[width=.33\textwidth]{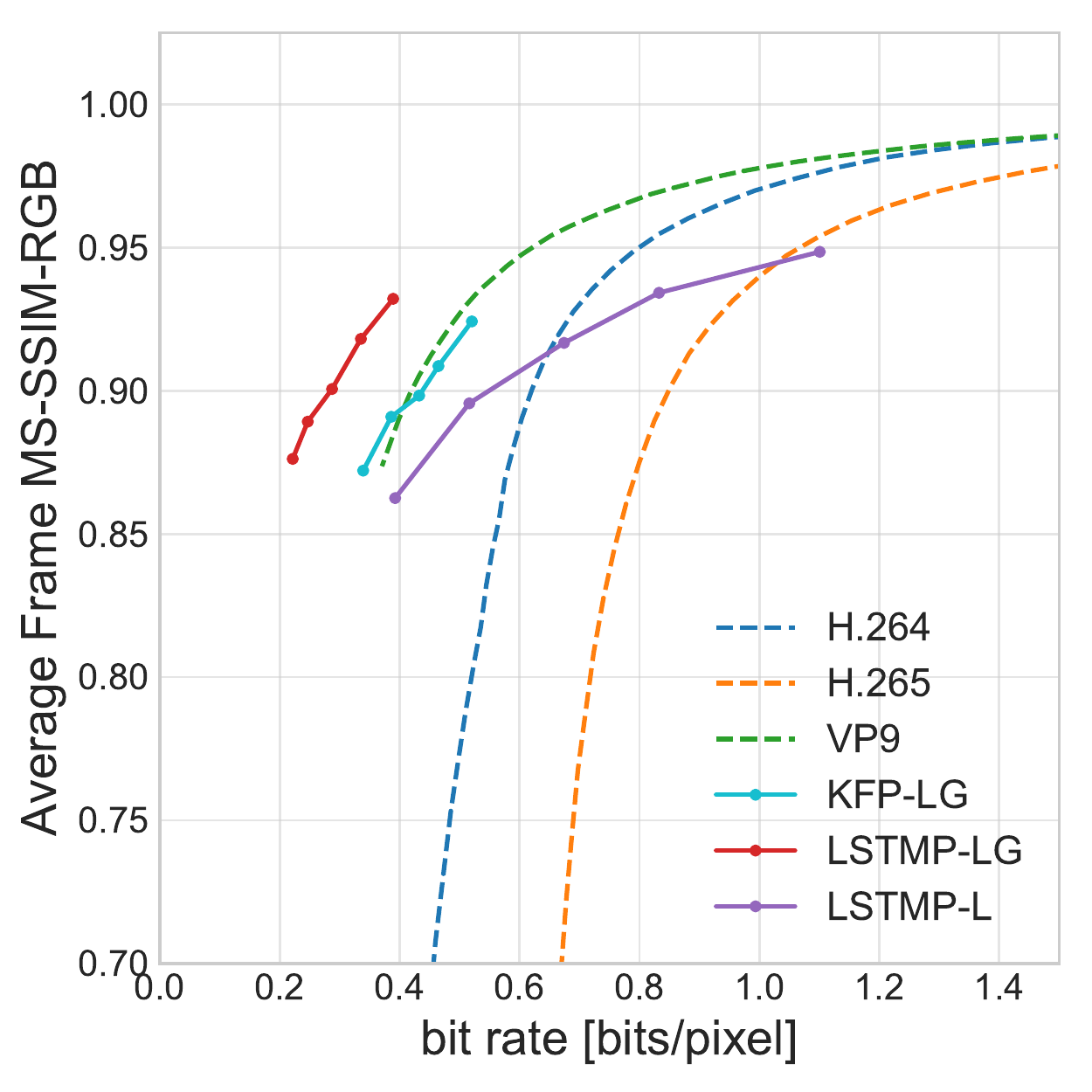}
\end{tabular}
\endgroup
\caption{\label{fig:msssim}Rate-distortion curves on three datasets measured in MS-SSIM (higher corresponds to lower distortion). Legend shared. Solid lines correspond to our models, with LSTMP-LG proposed.}
\end{figure*}

\textbf{MS-SSIM metric.} In the main paper, we evaluated performance in terms of PSNR distortion. Here, we also plot the MS-SSIM with respect to the bit rate to quantitatively compare our models to traditional codecs with respect to a perceptual metric. From Fig.~\ref{fig:msssim}, we can see that our LSTMP-LG saves significantly more bits when trained on specialized content videos and achieves competitive result when trained on general content videos.

\begin{figure}[t]
\begin{center}
\includegraphics[width=.45\textwidth]{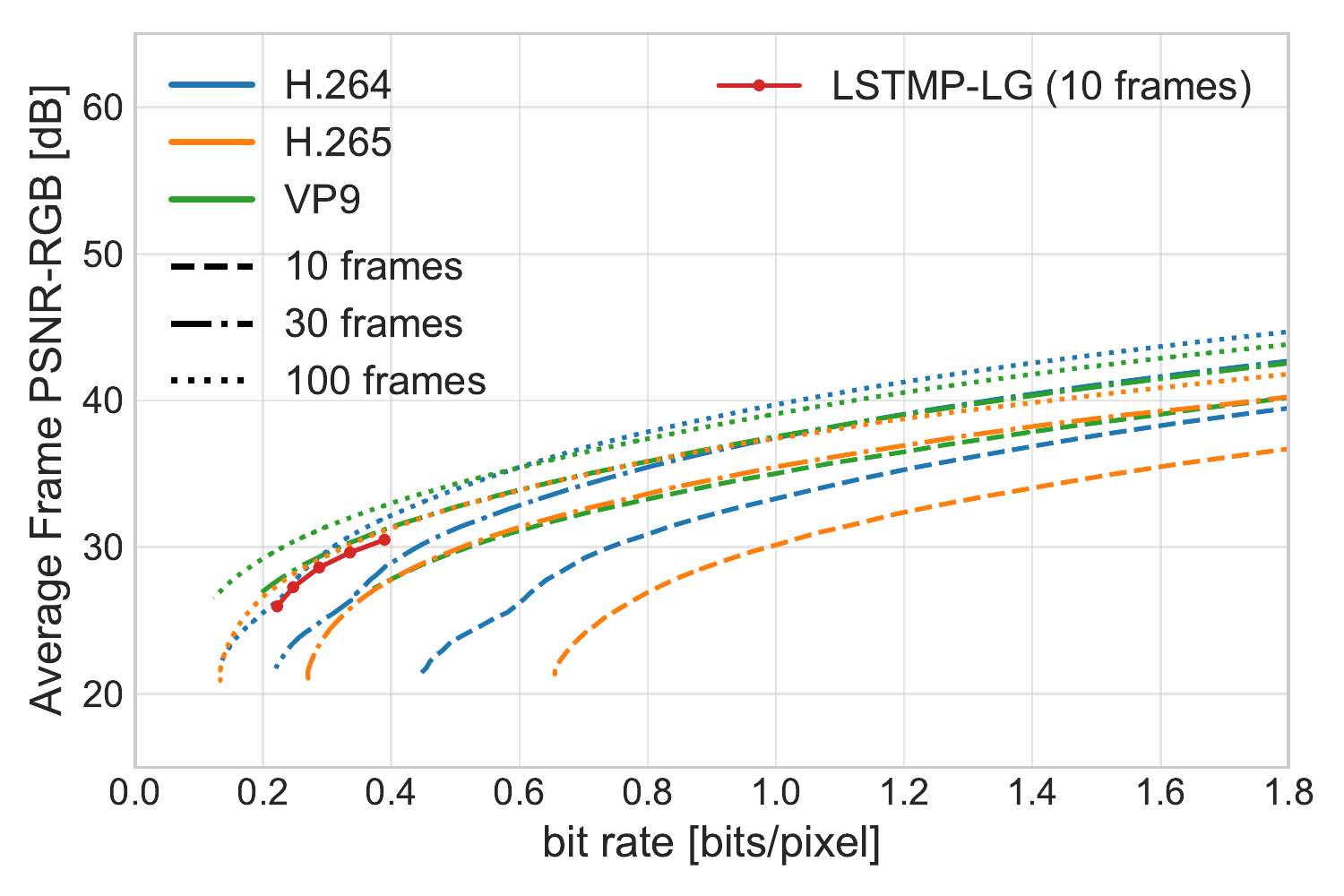}
\end{center}
\caption{\label{fig:PSNRvsT} Rate-distortion curves on the Kinetics dataset measured in PSNR. Codec performance is evaluated on video segments of $T = $ 10, 30, and 100 frames. Our best performing method (trained and evaluated on $T = 10$ frames) is shown in red for comparison.}
\end{figure}

\textbf{Longer videos.} We trained and evaluated our method on short video segments of $T=10$ frames and evaluated classical codec performance on the same segments. However, for typical videos, somewhat longer segments tend to have less information per pixel than $T=10$ segments, and standard video codecs are designed to take advantage of this fact. For this reason, we have presented video codec performances, evaluated on $T =$ 10, 30, and 100 frame segments for the Kinetics data in Fig.~\ref{fig:PSNRvsT}. While existing codec performance improves for longer segments, we note that our method (trained and evaluated on 10 frames) is still comparable to modern codec performance evaluated on longer segments. Additionally, with proper design and training on longer video segments, our method could be scaled to achieve similar temporal performance scaling since longer segments typically have less information per pixel. 

\end{document}

%% file: tex/math_commands.tex
\usepackage{amsmath,amsfonts,bm}

\def\eqref#1{equation~\ref{#1}}
\def\1{\bm{1}}

\def\vmu{{\bm{\mu}}}

\def\vf{{\bm{f}}}

\def\vv{{\bm{v}}}

\def\vx{{\bm{x}}}

\def\vz{{\bm{z}}}

\DeclareMathAlphabet{\mathsfit}{\encodingdefault}{\sfdefault}{m}{sl}
\SetMathAlphabet{\mathsfit}{bold}{\encodingdefault}{\sfdefault}{bx}{n}